\documentclass[twoside,11pt]{article}
\usepackage{jmlr2e} 

\jmlrheading{}{2021}{-}{2/21}{-/-}{-}{Goulet, Nguyen, and Amiri}

\ShortHeadings{Tractable Approximate Gaussian Inference for Bayesian Neural Networks}{Goulet, Nguyen, and Amiri}
\firstpageno{1}

\usepackage{microtype}
\usepackage{graphicx}
\usepackage{subfig}
\usepackage{booktabs} 
\usepackage{soul}
\usepackage[T1]{fontenc}
\usepackage[ttdefault=true]{AnonymousPro}
\usepackage{tikz}

\usepackage{bm}
\usepackage{multirow}
\usepackage{amsfonts,amssymb,latexsym,amscd,mathtools}       
\newcommand\independent{\protect\mathpalette{\protect\independenT}{\perp}}
\def\independenT#1#2{\mathrel{\rlap{$#1#2$}\mkern2mu{#1#2}}}

\usepackage{lastpage}
\jmlrheading{22}{2021}{1-\pageref{LastPage}}{9/20; Revised
6/21}{9/21}{20-1009}{James-A.~Goulet,~Luong Ha~Nguyen,~Saeid~Amiri}
\ShortHeadings{Tractable Approximate Gaussian Inference for Bayesian Neural Networks}{Goulet,~Nguyen,~Amiri}

\begin{document}

\title{Tractable Approximate Gaussian Inference for\\ Bayesian Neural Networks}

\author{\name James-A.~Goulet$^*$ \email james.goulet@polymtl.ca\\ 
\name Luong Ha~Nguyen\thanks{The first and second authors contributed equally to this work} \email luong-ha.nguyen@polymtl.ca\\ 
\name Saeid~Amiri\\
\addr Department of Civil Engineering, Polytechnique Montréal, Montréal, Canada}

\editor{Shakir Mohamed}

\maketitle

\begin{abstract}
In this paper, we propose an analytical method for performing \emph{tractable approximate Gaussian inference} (TAGI) in Bayesian neural networks. The method enables the analytical Gaussian inference of the posterior mean vector and diagonal covariance matrix for weights and biases. The method proposed has a computational complexity of $\mathcal{O}(n)$ with respect to the number of parameters $n$, and the tests performed on regression and classification benchmarks confirm that, for a same network architecture, it matches the performance of existing methods relying on gradient backpropagation. \end{abstract}


\section{Introduction}\label{S:INTRO}
The estimation of weight and bias in neural networks is currently dominated by approaches employing point estimates when learning model parameters using gradient backpropagation \citep{rumelhart1986learning}. Although these approaches allow for a state-of-the-art performance in many domains of applications, it is recognized that they fall short in situations where, for instance, datasets are small, when the task requires quantifying the uncertainty about the prediction made, and for continual learning \citep{ghahramani2015probabilistic,kendall2017uncertainties,farquhar2019unifying}. In general, the Bayesian approach for inferring the parameters'\! posterior is known to be theoretically better suited than a point estimate \citep{goodfellow2016deep,murphy2012machine}. This is in theory because applying exact Bayesian inference on large neural networks has been considered to be intractable \citep{goodfellow2016deep}. 

In this paper, we propose a \emph{tractable approximate Gaussian inference} method (TAGI) for Bayesian neural networks.  The approach, which does not rely on backpropagation, allows for an analytical treatment of uncertainty for the weight and bias parameters. TAGI relies on two key steps: The first one consists in the analytical forward uncertainty propagation through the network using moment generating functions and a local linearization of activation functions. This step allows computing the expected values, covariances, and cross covariances required for performing analytical inference. The second step leverages the Gaussian assumptions throughout the network in order to perform the analytical Gaussian inference for both the hidden units and parameters. The key to the computational tractability of TAGI is to take advantage of the inherent conditional independence embedded in neural network in order to perform both the forward propagation of uncertainty as well as the Gaussian inference in a recursive layer-wise manner. This allows reaching a storage as well as the computational complexity that is linear with respect to the number of weights in the network.  

 The paper is organized as follows: Section \ref{S:GCBNN} first introduces the \emph{Gaussian multiplication approximation} (GMA) for propagating uncertainty in feedforward neural networks, and second, it presents how to perform tractable Gaussian inference for the posterior mean vector and diagonal covariance for the weight and bias parameters. Section \ref{S:related_work}, positions TAGI in the context of the related work which has already tackled the problem of approximate inference in Bayesian neural networks. Finally, Section \ref{S:experiments} validates the performance of the approach on benchmark regression problems and on the MNIST classification problem. 
\section{Gaussian Approximation for BNN}\label{S:GCBNN}
This section first introduces the nomenclature for Gaussian feedforward neural network, then we present how to use matrix operations in order to propagate uncertainty through it, and finally, we present how to perform tractable approximate Gaussian inference.
\subsection{Gaussian Feedforward Neural Network}\label{SS:FFNN}
Let us consider a vector of input covariates $\bm{X}=[X_{1}~X_{2}~\ldots~X_{\mathtt{X}}]^{\intercal}$ such that $\bm{x}\in\mathbb{R}^{{\mathtt{X}}}$ that are described by random variables in order to take into account errors potentially arising from observation uncertainties, and then suppose we have a vector of $\mathtt{Y}$ observed system responses $\bm{Y}=[Y_{1}~Y_{2}~\ldots~Y_{\mathtt{Y}}]^{\intercal}$ such that $\bm{y}\in\mathbb{R}^{{\mathtt{Y}}}$. Note that throughout the paper, the number of variables in vectors or sets are denoted by typewriter typefaces such as $\mathtt{X}$ and $\mathtt{Y}$. The details of the nomenclature for the feedforward neural network (FNN) employed in this paper is presented in Figure \ref{FIG:FNN} from Appendix \ref{A:FNN}. The relations between observed system responses and its covariates are described by the observation model
\begin{equation}
\bm{y}=\bm{z}^{(\mathtt{O})}+\bm{v},
\label{EQ:obs}
\end{equation}
where the vector of hidden variables $\bm{z}^{(\mathtt{O})}$ corresponds to the output layer of a neural network on which observation errors $\bm{v}$ are added such that $\bm{V}\sim\mathcal{N}(\bm{v};\mathbf{0},\mathbf{\Sigma}_{\bm{V}})$. In common cases, $\mathbf{\Sigma}_{\bm{V}}$ is a diagonal covariance matrix assuming that observation errors are independent from each other, and further, we assume that $\mathbf{\Sigma}_{\bm{V}}$ is independent of $\bm{x}$. The treatment heteroschedastic problems with a closed-form analytical inference for $\sigma_V(x)$ is outside the scope of this paper and is not further treated. We can model the relations between covariates $\bm{x}$ and output hidden variables $\bm{z}^{(\mathtt{O})}$ using a feedforward neural network consisting of $\mathtt{L}$ hidden layers each having $\mathtt{A}$ activation units $a^{(j)}_{i}$ and hidden variables $z^{(j)}_{i}$, $\forall i=\{1,2,\cdots,\mathtt{A}\}$, where an activation unit $a^{(j)}_{i}$ is a non-linear transformation of its associated hidden variable, $a_{i}^{(j)}=\sigma(z_{i}^{(j)})$. Note that for the sake of simplifying the notation and explanations, throughout the paper, we consider that all hidden layers have the same number of units $\mathtt{A}$.     
We go from the input layer containing the covariates $\bm{x}$, to the $i^{\text{th}}$ hidden variable on the first hidden layer, using an affine function of $\bm{x}$ so that
\begin{equation}
z_{i}^{(1)}=w_{i,1}^{(\texttt{0})}x_{1}+w_{i,2}^{(\texttt{0})}x_{2}+\cdots+w_{i,\mathtt{X}}^{(\texttt{0})}x_{\mathtt{X}}+b^{(\texttt{0})}_{i}.
\label{E:input_layer}
\end{equation}
Equation \ref{E:input_layer} involves the product of weights $w_{i,j}^{(\texttt{0})}$ and covariates $x_{j}$ with an additive bias term $b^{(\texttt{0})}_{i}$. In the context of a neural network, the process of learning consists in estimating these weights and biases. Here, we consider that weight and bias parameters are described by random variables so that our joint prior for $\{\bm{X},\bm{W}^{(\texttt{0})},\bm{B}^{(\texttt{0})}\}$ is a multivariate Gaussian. Although we know this distribution assumption may not be suited in all situations, we restrict ourself to the Gaussian setup.


In Equation \ref{E:input_layer}, we note that the product of Gaussian random variables is not Gaussian; Despite this, we propose to employ moment generating functions in order to compute analytically its expected value, its variance, as well as the covariance between the product of Gaussian random variables and any other Gaussian random variable. 

For instance, let $\bm{X}=[X_1~\ldots~X_4]^{\intercal}\sim\mathcal{N}(\bm{x};\bm{\mu},\mathbf{\Sigma})$ be a generic vector of Gaussian random variables, where $\bm{\mu}$ is the mean vector and $\mathbf{\Sigma}$ is the covariance matrix, then the following statements hold,  \vspace{-4mm}
\begin{eqnarray}
\mathbb{E}[X_1X_2]\!\!\!\!\!&=&\!\!\!\!\!\mu_1\mu_2+\text{cov}(X_1,X_2), \label{eq1}\\[4pt]
\text{cov}(X_3,X_1X_2)\!\!\!\!\!&=&\!\!\!\!\!\text{cov}(X_1,\!X_3)\mu_2\!+\!\text{cov}(X_2,\!X_3)\mu_1 \label{eq2},\\[4pt]
\text{cov}(X_1X_2,X_3X_4)\!\!\!\!\!&=&\!\!\!\!\!\text{cov}(X_1,X_3)\text{cov}(X_2,X_4) +\text{cov}(X_1,X_4)\text{cov}(X_2,X_3)\label{eq3}   \\[2pt]
\!\!\!\!\!&&\!\!\!\!\!\!\!\!\!\!\!\!\!\!\!\!\!\!+\text{cov}(X_1,X_3)\mu_2\mu_4+\text{cov}(X_1,X_4)\mu_2\mu_3\nonumber  \\
\!\!\!\!\!&&\!\!\!\!\!\!\!\!\!\!\!\!\!\!\!\!\!\!+\text{cov}(X_2,X_3)\mu_1\mu_4+\text{cov}(X_2,X_4)\mu_1\mu_3, \nonumber\\[4pt]
\text{var}(X_1X_2)\!\!\!&=&\!\!\! \sigma_{1}^{2}\sigma_{2}^{2}+\text{cov}(X_1,X_2)^{2}+2\text{cov}(X_1,X_2)\mu_{1}\mu_{2}+\sigma_{1}^{2}\mu_{2}^{2}+\sigma_{2}^{2}\mu_{1}^{2}. \label{eq4}
\end{eqnarray}
The development of statements (\ref{eq1}--\ref{eq4}) is presented in the Appendix \ref{Appendix:A}. In this paper, we define the \emph{Gaussian multiplication approximation} as the approximation of the probability density function (PDF) for any product term  $X_{i}X_{j}$ by a Gaussian whose first two moments are defined by Equations (\ref{eq1}--\ref{eq4}). With this approximation, we can now employ $X_{i}X_{j}$ along with the random state vector ${\bm{X}}$ in affine functions. This allows propagating the uncertainty from the input covariates and prior knowledge on weight and bias parameter through a FNN. 

\begin{figure}[b!]\vspace{-4mm}
\subfloat[][$\bm{a}\to{z}^{+}_1$]{
\raisebox{4mm}{\!\!\!\!\!\!\!\!\!\!\!\!\!\!\!\!\!\!\!\!\!\!\input{FeedForward_NN_layer_GMA.tex}}}~
\subfloat[][$\mathtt{A}=1$]{\includegraphics[width=40mm]{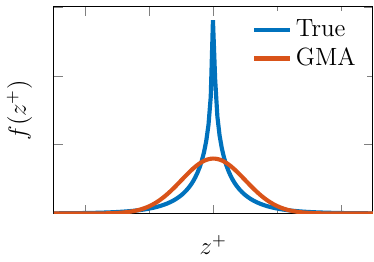}}~
\subfloat[][$\mathtt{A}=5$]{\includegraphics[width=40mm]{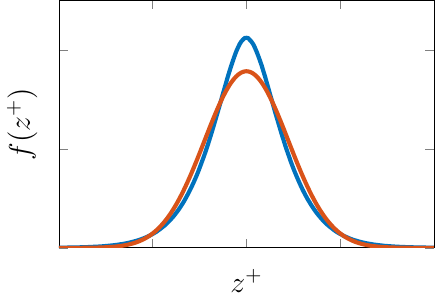}}~
\subfloat[][$\mathtt{A}=10$]{\includegraphics[width=40mm]{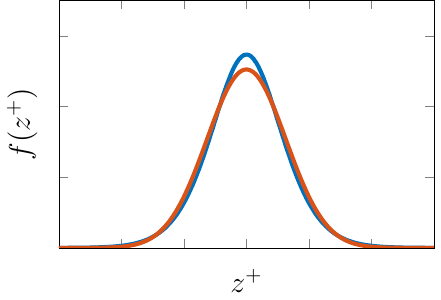}}
\caption{Illustration of the effect of the GMA on the PDF of a hidden unit ${Z}^{+}_i$ as a function of the number of activation units $\mathtt{A}$ on the preceding hidden layer. The blue curve represents the true PDF estimated using sampling and the red one, the PDF resulting from the GMA.}
\label{FIG:GMA_CLT}
\end{figure}

\begin{figure}[h!]\centering
\includegraphics[width=0.45 \columnwidth]{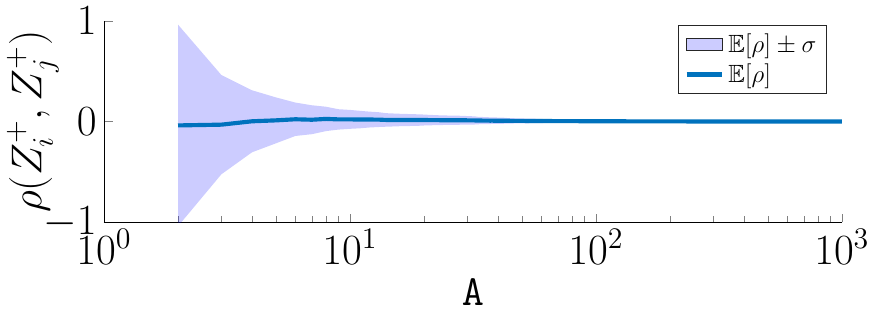}\\[-6pt]
\caption{Evolution of the correlation $\rho(Z^+_i,Z^+_j)$ as a function of the number of activation units $\mathtt{A}$ on the preceding layer. The expected value and the confidence region corresponding to one standard deviation are presented for 1000 random realizations of weight values.}
\label{FIG:Gaussian_conditional_rho_prior}
\end{figure}

Figure \ref{FIG:GMA_CLT}a illustrates the passage from the activation units $\bm{A}$, to a subsequent hidden unit $\bm{Z}^{+}_i$. Figures \ref{FIG:GMA_CLT}b--d compare the true theoretical PDFs with those obtained using the GMA for different numbers of activation units $\mathtt{A}$, under the assumption that both $A_k\sim\mathcal{N}(a_k;0,1)$, $W_{i,k}\sim\mathcal{N}(w_{i,k};0,1)$ and $b_i=0$. This shows that even if the GMA is a crude approximation for the true PDF resulting from the product of two Gaussians, when several of these independent product terms are added, the result quickly tends to a Gaussian-like PDF. In order for the central limit theorem (CLT) to apply, all activation units $A_k$ must be independent. We can demonstrate the validity of this hypothesis by revisiting the same example while computing the correlation between pairs of hidden units $\{{Z}^{+}_i,{Z}^{+}_j\}$ on the subsequent hidden layer. With this example, we want to show that starting from a hidden layer having $\mathtt{A}$ independent activation units, pairs of hidden units on the subsequent layer are also independent if $\mathtt{A}$ is sufficiently large, despite the fact that the hidden units $\{{Z}^{+}_i,{Z}^{+}_j\}$ depend on the same activation units on the previous layer. Figure \ref{FIG:Gaussian_conditional_rho_prior} presents the expected value, and the one standard deviation confidence region for $\rho(Z^+_i,Z^+_j)$, obtained from 1000 weight value realizations. The wide confidence region associated with small numbers of units, e.g. $\mathtt{A}<10$, confirms that for most realizations, the dependence on the same activation units will introduce a strong correlation between the hidden units of the subsequent layer. On the other hand, for larger $\mathtt{A}$ values, the narrow confidence region centred around $0$ confirms that for any sets of weight realizations, the hidden units on the subsequent layer remain independent. This shows that starting from independent activation units, any pairs of hidden units on the following layer are also approximately independent.  Because neural networks employ the same operations recursively over multiple layers, this independence assumption for the hidden units remains valid throughout the network. The theoretical example presented here is a simplified version of a real neural network as weights are randomly sampled;  For real neural networks, \cite{DBLP:confWuNMTHG19} have confirmed empirically that some form of CLT hold for the hidden units during training as the independence hypothesis remained approximately valid between layers. In Section \ref{SS:toy_problem} we further demonstrate on a real problem that if $\mathtt{A}$ is large enough, the approximate independence of the hidden units within layers remains valid. 

\begin{figure}[t]
\centering
\includegraphics[width=55mm]{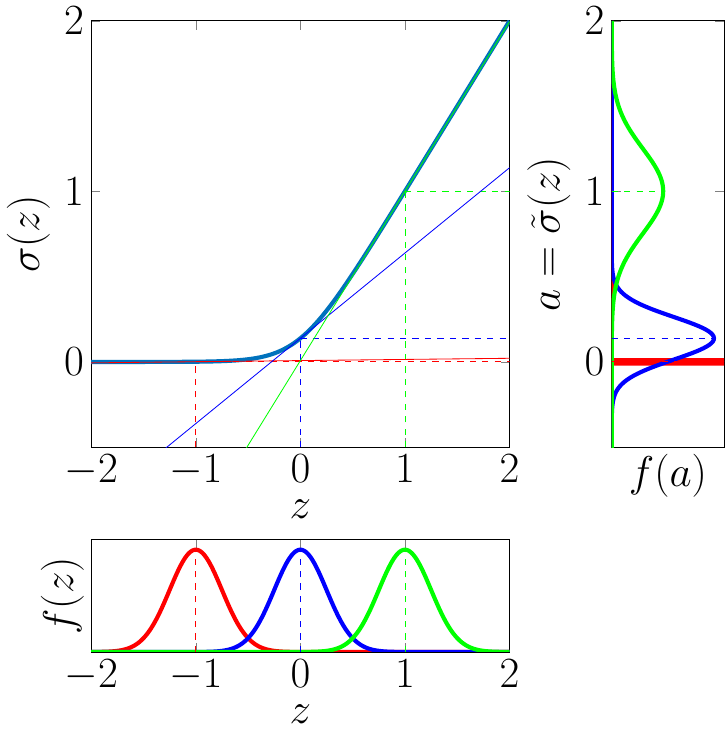}\\[-4pt]
\caption{Examples of local linearization for a softplus activation function ${\sigma}(\cdot)$ at the expected value $\mathbb{E}[{Z}]$ for $Z\sim\mathcal{N}(z;\mu_Z,0.25^2)$ where $\mu_Z=\{-1, 0, 1\}$.}
\label{FIG:LACT}
\end{figure}

The passage from the hidden variables $\bm{Z}$, to their corresponding activation units $\bm{A}$ cannot be done analytically using non-linear activation functions. In order to work around this difficulty, we propose to employ functions that are locally linearized at $\mathbb{E}[\bm{Z}]=\bm{\mu_Z}$, analogously to what is done for the extended Kalman filter \citep{haykin2004kalman}. \emph{Locally linearized activation functions} $\tilde{\sigma}(\cdot)$ allow calculating analytically the expected vector $\mathbb{E}[\bm{A}]$, the covariance $\text{cov}(\bm{A})$, as well as the cross-covariance between activation units and the weight and bias $\text{cov}(\bm{A},\bm{\theta})$, where $\bm{\theta}=\{\bm{W},\bm{B}\}$. Figure \ref{FIG:LACT} presents an example for the linearization of a softplus activation function. Note that a locally linearized activation function is not equivalent to having a linear activation function because for each instance of input covariates $\bm{x}_i$, the linearization is done at a different value $\mu_z$, which maintains the non-linear dependency between the input $\bm{x}_i$ and the output $\bm{y}_i$. This linearization procedure is an approximation of the \emph{change of variable rule} (\citeauthor{murphy2012machine}, 2012, \S2.6.2) that would be required to obtain the true theoretical PDF for the output $f(a)$ from the input $f(z)$. Although it is known that this approach may lead to poor approximation \citep{wan2000unscented} when there is a  non-linearity in a region high probability density, we choose to employ this linearization procedure because of its minimal computational cost which still allows, as it will be shown in Section \ref{S:experiments}, matching the state-of-the-art performance on equivalent neural networks architectures using backpropagation. Finally, note that the linearization procedure is compatible with all the common activation functions such as the ReLU, tanh, logistic sigmoid, etc. 

The transition from the knowledge of the $j^{\text{th}}$ layer's activation units to an activation unit on the $j+1$ layer is defined by $a_{i}^{(j+1)}=\tilde{\sigma}(z_{i}^{(j+1)})=\tilde{\sigma}(w_{i,1}^{(j)}a_{1}^{(j)}+w_{i,2}^{(j)}a_{2}^{(j)}+\cdots+w_{i,\mathtt{A}}^{(j)}a_{\mathtt{A}}^{(j)}+b^{(j)}_{i})$. Analogously, we go from the last hidden layer to the output layer by following $$z_{i}^{(\mathtt{O})}=w_{i,1}^{(\mathtt{L})}a_{1}^{(\mathtt{L})}+w_{i,2}^{(\mathtt{L})}a_{2}^{(\mathtt{L})}+\cdots+w_{i,\mathtt{A}}^{(\mathtt{L})}a_{\mathtt{A}}^{(\mathtt{L})}+b^{(\mathtt{L})}_{i},$$
for all $i\in\{1,2,\cdots,\mathtt{Y}\}$. All these steps define what we call the \emph{approximate Gaussian feedforward neural network} (AG-FNN), which can be summarized by the input-output relation \begin{equation}\!\left\{\bm{\mu}_{\bm{Z}}^{(\mathtt{O})},\mathbf{\Sigma}_{\bm{Z}}^{(\mathtt{O})},\mathbf{\Sigma}_{\bm{Z\theta}}^{(\mathtt{O})}\right\}=\text{AG-FNN}\left(\bm{\mu}_{\bm{X}},\mathbf{\Sigma}_{\bm{X}},\bm{\mu}_{\bm{\theta}},\mathbf{\Sigma}_{\bm{\theta}}\right).\label{EQ:LFNN}\end{equation} 

For the \emph{regression setup} where $\bm{y}\in\mathbb{R}^{\mathtt{\mathtt{Y}}}$, the observed model output is directly defined by the last layer as described in Equation \ref{EQ:obs} so that $\mathcal{N}(\bm{y};\bm{\mu}_{\bm{Y}},\mathbf{\Sigma}_{\bm{Y}})$, where, $\bm{\mu}_{\bm{Y}}=\bm{\mu}_{\bm{Z}}^{(\mathtt{O})}$, and $\mathbf{\Sigma}_{\bm{Y}}=\mathbf{\Sigma}_{\bm{Z}}^{(\mathtt{O})}+\mathbf{\Sigma}_{\bm{V}}$. 

For the \emph{classification setup}, we need to convert the output $\bm{y}\in\mathbb{R}^{\mathtt{\mathtt{Y}}}$ into a class observation ${y}^{(\mathtt{C})}\in\{1,2,\cdots,\mathtt{K}\}$. Note that using the traditional \emph{Softmax} output layer would not allow for a closed-form solution for propagating and marginalizing the uncertainty associated with the output $\bm{Y}$. Instead, we propose to employ a \emph{hierarchical binary decomposition} similar to what was employed by \citeauthor{morin2005hierarchical} (\citeyear{morin2005hierarchical}). The details regarding this formulation are presented in Appendix \ref{A:binary_tree}.

\subsection{Matrix operations for AG-FNN}\label{SS:LAFFNN}
In this section, we describe how the steps involved in the evaluation of Equation \ref{EQ:LFNN} can be performed using matrix operations. Our first hypothesis supposes that the knowledge for covariates, hidden units, as well as the weights and biases, is described by Gaussian random variables. We can then generalize the operations for going from the $\mathtt{A}$ activation units at a layer $j$ to the subsequent $\mathtt{A}$ hidden units at layer $j+1$ using matrix operations so that
\begin{equation}\begingroup 
\setlength\arraycolsep{2pt}
\underbrace{\begin{bmatrix}Z_{1}^{(j+1)}\\Z_{2}^{(j+1)}\\\vdots\\Z_{\mathtt{A}}^{(j+1)}\end{bmatrix}}_{\bm{Z}^{(j+1)}}\!=\!\underbrace{\begin{bmatrix}W_{1,1}^{(j)}&W_{1,2}^{(j)}&\cdots&W_{1,\mathtt{A}}^{(j)}\\W_{2,1}^{(j)}&W_{2,2}^{(j)}&\cdots&W_{2,\mathtt{A}}^{(j)}\\\vdots&\vdots&\ddots&\vdots\\W_{\mathtt{A},1}^{(j)}&W_{\mathtt{A},2}^{(j)}&\cdots&W_{\mathtt{A},\mathtt{A}}^{(j)}
\\\end{bmatrix}}_{\bm{W}^{(j)}}\!\times\!\underbrace{\begin{bmatrix}A_{1}^{(j)}\\A_{2}^{(j)}\\\vdots\\A_{\mathtt{A}}^{(j)}\end{bmatrix}}_{\bm{A}^{(j)}}\!+\!\underbrace{\begin{bmatrix}B^{(j)}_{1}\\B^{(j)}_{2}\\\vdots\\B^{(j)}_{\mathtt{A}}\end{bmatrix}}_{\bm{B}^{(j)}}\label{E:Z_matrix}.
\endgroup\end{equation}
Our prior knowledge for activation units is described by $\bm{A}^{(j)}\sim\mathcal{N}(\bm{a}^{(j)};\bm{\mu}^{(j)}_{\bm{A}},\mathbf{\Sigma}^{(j)}_{\bm{A}})$ as well as by the cross-covariance $\text{cov}(\bm{\theta},\bm{A}^{(j)})$ between the activation units and the vector $\bm{\theta}\in\mathbb{R}^{\mathtt{P}}$ containing all the weight and bias parameters defined for all the layers in the network. We can re-write Equation \ref{E:Z_matrix} by breaking down the matrix-vector product $\bm{W}\times\bm{A}$, into an operation-wise equivalent vector $(\bm{W\!\!A})\in\mathbb{R}^{\mathtt{A}^2\times1}$ for which the moments of the product terms can be pre-computed; We can employ Equation \ref{eq1} in order to compute the expected vector $\bm{\mu}^{(j)}_{\bm{W\!\!A}}\equiv\mathbb{E}[(\bm{W\!\!A})^{(j)}]$, Equation \ref{eq3}--\ref{eq4} for the covariance matrix $\mathbf{\Sigma}^{(j)}_{\bm{W\!\!A}}\equiv\text{cov}\left((\bm{W\!\!A})^{(j)}\right)\in\mathbb{R}^{\mathtt{A}^{2}\times \mathtt{A}^{2}}$ associated with the product terms $(\bm{W\!\!A})^{(j)}\in\mathbb{R}^{\mathtt{A}^{2}}$, and Equation \ref{eq2} for the cross-covariance matrix $\mathbf{\Sigma}^{(j)}_{\bm{W\!\!A\theta}}\equiv\text{cov}\left((\bm{W\!\!A})^{(j)},\bm{\theta}\right)\in\mathbb{R}^{\mathtt{A}^{2}\times \mathtt{P}}$. We then introduce two new deterministic matrices $\mathbf{F}_{\bm{wa}}^{(j)}\in\{0,1\}^{\mathtt{A}\times\mathtt{A}^{2}}$ and $\mathbf{F}_{\bm{b}}^{(j)}\in\{0,1\}^{\mathtt{A}\times\mathtt{A}}$, which allow rewriting Equation \ref{E:Z_matrix} as a system of linear equations involving the product-terms vector $(\bm{W\!\!A})$, 
\begin{equation}\bm{Z}^{(j+1)}=\mathbf{F}_{\bm{wa}}^{(j)} (\bm{W\!\!A})^{(j)}+\mathbf{F}_{\bm{b}}^{(j)}\bm{B}^{(j)}.\label{E:layer_matrix}\end{equation}
Note that $\mathbf{F}_{\bm{wa}}^{(j)}$ and $\mathbf{F}_{\bm{b}}^{(j)}$ are non-unique as their specific definition depend on the ordering of variables in the problem. An example of structure for these matrices is presented in Appendix \ref{Appendix:B}. Using the properties for linear functions of Gaussian random variables, we obtain
\begin{equation}
\!\!\!\!\!\begin{array}{rrcl}
\bm{\mu}^{(j+1)}_{\bm{Z}}\equiv&\!\!\!\!\!\mathbb{E}[\bm{Z}^{(j+1)}]&\!\!\!\!\!=\!\!\!\!\!&\mathbf{F}_{\bm{wa}}^{(j)}\bm{\mu}^{(j)}_{\bm{W\!\!A}}\!\!+\!\mathbf{F}_{\bm{b}}^{(j)}\!\bm{\mu}^{(j)}_{\bm{B}},\\[4pt]
\mathbf{\Sigma}^{(j+1)}_{\bm{Z}}\equiv&\!\!\!\!\!\text{cov}(\bm{Z}^{(j+1)})&\!\!\!\!\!=\!\!\!\!\!&\mathbf{F}_{\bm{wa}}^{(j)} \mathbf{\Sigma}^{(j)}_{\bm{W\!\!A}}\mathbf{F}_{\bm{wa}}^{(j)\intercal}+\mathbf{F}_{\bm{b}}^{(j)}\mathbf{\Sigma}^{(j)}_{\bm{B}}\mathbf{F}_{\bm{b}}^{(j)\intercal}+2\mathbf{F}_{\bm{wa}}^{(j)}\text{cov}(\bm{W\!\!A}^{(j)}\!,\!\bm{B}^{(j)})\mathbf{F}_{\bm{b}}^{(j)\intercal},\\[4pt]
\mathbf{\Sigma}^{(j+1)}_{\bm{Z\theta}}\equiv&\!\!\!\!\text{cov}(\bm{Z}^{(j+1)}\!\!,\bm{\theta})&\!\!\!\!\!=\!\!\!\!\!&\mathbf{F}_{\bm{wa}}^{(j)}\mathbf{\Sigma}^{(j)}_{\bm{W\!\!A\theta}}+\mathbf{\Sigma}^{(j)}_{\bm{B\theta}},\label{E:z_covariance}
\end{array}
\end{equation}
where $\mathbf{\Sigma}^{(j)}_{\bm{B\theta}}\equiv \text{cov}(\bm{B}^{(j)},\bm{\theta})\in\mathbb{R}^{\mathtt{A}\times \mathtt{P}}$ is the covariance between the bias parameters from the $j^{\text{th}}$ layer and all the other parameters. In order to apply the locally linearized activation function $\bm{A}^{(j+1)}=\tilde{\sigma}(\bm{Z}^{(j+1)})$,
\begin{equation}\bm{A}^{(j+1)}=\mathbf{J}^{(j+1)}\left(\bm{Z}^{(j+1)}-\bm{\mu}^{(j+1)}_{\bm{Z}}\right)+{\sigma}(\bm{\mu}^{(j+1)}_{\bm{Z}}),\label{EQ:activation}
\end{equation} we need to define the diagonal Jacobian matrix of the transformation evaluated at $\bm{\mu}^{(j+1)}_{\bm{Z}}$, 
$\mathbf{J}^{(j+1)}=\text{diag}\big(\nabla_{\!\!\bm{z}} {\sigma}(\bm{\mu}^{(j+1)}_{\bm{Z}})\big)$.
Using again the properties for linear functions of Gaussian random variables, we obtain
\begin{equation}
\begin{array}{rrcl}
\bm{\mu}^{(j+1)}_{\bm{A}}\equiv&\mathbb{E}[\bm{A}^{(j+1)}]&=&\tilde{\sigma}(\bm{\mu}^{(j+1)}_{\bm{Z}}),\\[4pt]
\mathbf{\Sigma}^{(j+1)}_{\bm{A}}\equiv&\text{cov}(\bm{A}^{(j+1)})&=&\mathbf{J}^{(j+1)}\mathbf{\Sigma}^{(j+1)}_{\bm{Z}}\mathbf{J}^{(j+1)\intercal},\\[4pt]
\mathbf{\Sigma}^{(j+1)}_{\bm{A\theta}}\equiv&\text{cov}(\bm{A}^{(j+1)},\bm{\theta})&=&\mathbf{J}^{(j+1)}\mathbf{\Sigma}^{(j+1)}_{\bm{Z\theta}}.\label{E:a_covariance}
\end{array}
\end{equation}
Equations \ref{E:z_covariance} \& \ref{E:a_covariance} allow propagating the information about the covariance of activation units and its dependence on parameters through any pairs of successive layers. For the input layer, the steps described in Equations \ref{E:Z_matrix}--\ref{E:a_covariance} remain the same except that the activation units $\bm{A}^{(j)}$ are replaced by the covariates $\bm{X}\sim\mathcal{N}(\bm{x};\bm{\mu}_{\bm{X}},\mathbf{\Sigma}_{\bm{X}})$. 

\subsection{Tractable Approximate Gaussian Inference (TAGI)}\label{SS:bayesian_estimation}
Let us assume  we have a set of joint observations for covariates and system responses so that $\mathcal{D}=\{\mathcal{D}_{x},\mathcal{D}_{y}\}=\{(\bm{x}_{i},\bm{y}_{i}),\forall i\in\{1:\mathtt{D}\}\}$. Given that our prior knowledge for the neural network's parameter is $f(\bm{\theta})=\mathcal{N}(\bm{\theta};\bm{\mu}_{\bm{\theta}},\mathbf{\Sigma}_{\bm{\theta}})$, the method presented in \S\ref{SS:FFNN} supposes that the joint PDF $f(\bm{\theta},\bm{y})$ for parameters $\bm{\theta}$ and observations $\bm{y}$ is Gaussian with mean vector and covariance
$$\bm{\mu}=\left[\begin{array}{c}\bm{\mu}_{\bm{\theta}}\\\bm{\mu}_{\bm{Y}}\end{array}\right] ,~ \mathbf{\Sigma}=\left[\begin{array}{cc}\mathbf{\Sigma}_{\bm{\theta}}& \mathbf{\Sigma}_{\bm{Y\!\theta}}^\intercal\\
\mathbf{\Sigma}_{\bm{Y\!\theta}}& \mathbf{\Sigma}_{\bm{Y}} 
\end{array}\right].$$
The Gaussian inference for the vector $\bm{\theta}$ given observations $\bm{Y}=\bm{y}$ is described by the Gaussian conditional equations $f(\bm{\theta}|\bm{y})=\mathcal{N}(\bm{\theta};\bm{\mu}_{\bm{\theta}|\bm{y}},\mathbf{\Sigma}_{\bm{\theta}|\bm{y}})$ defined by its conditional mean vector and covariance matrix,
\begin{equation}\begin{array}{l}
\bm{\mu}_{\bm{\theta}|\bm{y}}=\bm{\mu}_{\bm{\theta}}+\mathbf{\Sigma}_{\bm{Y\!\theta}}^{\intercal}\mathbf{\Sigma}_{\bm{Y}}^{-1}(\bm{y}-\bm{\mu}_{\bm{Y}})\\[4pt]
\mathbf{\Sigma}_{\bm{\theta}|\bm{y}}=\mathbf{\Sigma}_{\bm{\theta}}-\mathbf{\Sigma}_{\bm{Y\!\theta}}^{\intercal}\mathbf{\Sigma}_{\bm{Y}}^{-1}\mathbf{\Sigma}_{\bm{Y\!\theta}}.\end{array}\label{E:Gaussian_conditional_theta}\end{equation}
This 1-step network-wise Gaussian inference procedure is computationally prohibitive because the forward propagation of uncertainty depicted in Figure \ref{FIG:FNN_inference_long}a involves large-sized densely populated matrices, and the inference using Equation \ref{E:Gaussian_conditional_theta} again involves full matrices.
{\captionsetup[subfloat]{farskip=-5pt,captionskip=-3pt}
\begin{figure}[h!]
\!\!\!\!\!\!\!\!\!\!\!\!\!\!\!\!\!\!\subfloat[][Intractable 1-step network-wise inference (Eq. \ref{E:Gaussian_conditional_theta})]{\!\!\!\!\!\!\!\!\!\!\!\!\tikzset{dist1/.style={path picture= {
    \begin{scope}[x=1pt,y=10pt]
      \draw plot[domain=-6:6] (\x,{1/(1 + exp(-\x))-0.5});
    \end{scope}
    }
  }
}
\tikzset{dist2/.style={path picture= {
    \begin{scope}[x=1pt,y=10pt]
      \draw plot[domain=-6:6] (\x,{\x/10});
    \end{scope}
    }
  }
}
\tikzstyle{input}=[draw,fill=red!10,circle,minimum size=15pt,inner sep=0pt]
\tikzstyle{hidden}=[draw,fill=green!20,circle,minimum size=15pt,inner sep=0pt]
\tikzstyle{output}=[draw,fill=blue!20,circle,minimum size=15pt,inner sep=0pt]
\tikzstyle{bias}=[draw=none,circle,minimum size=15pt,inner sep=0pt]

\tikzstyle{stateTransition}=[->,line width=0.25pt,draw=black!50]
\tikzstyle{Forward}=[->,line width=0.25pt,draw=magenta!75]
\tikzstyle{Inference}=[->,line width=0.25pt,draw=cyan!75]

\begin{tikzpicture}[scale=1.45]\scriptsize
    \node (x)[input]   at (0, 0) {$\bm{x}$};
    \node (z1)[hidden] at (1, 0) {$\,\bm{z}^{\!(1)}$};
    \node (zi) at (2, 0) {$\cdots$};
    \node (zL)[hidden] at (3, 0) {$\,\bm{z}^{\!(\mathtt{L})}$};
    \node (zO)[hidden] at (4,0) {$\,\bm{z}^{\!(\mathtt{O})}$};
    \node (y)[output] at (4.85,0) {$\bm{y}$};
    
    \node (xp)[minimum size=10pt]   at (0.5, 0) {};
    \node (z1p)[minimum size=10pt] at (1.5, 0) {};
    \node (zip)[minimum size=8pt] at (2.5, 0) {};
    \node (zLp)[minimum size=7pt] at (3.5, 0) {};

    \draw[stateTransition,opacity=1] (x) -- (z1) node [midway, rotate=0,fill=white,opacity=0.8] {\scriptsize$\!\bm{\theta}^{(0)}\!$};
    \draw[stateTransition,opacity=1] (z1) -- (zi) node [midway, rotate=0,fill=white,opacity=0.8] {\scriptsize$\!\bm{\theta}^{(1)}\!$};
          \draw[stateTransition,opacity=1] (zi) -- (zL) node [midway, rotate=0,fill=white,opacity=0.8] {\scriptsize$\!\!\bm{\theta}^{(\mathtt{L}\text{-}1)}\!\!$};
    \draw[stateTransition,opacity=1] (zL) -- (zO) node [midway, rotate=0,fill=white,opacity=0.8] {\scriptsize$\!\bm{\theta}^{(\mathtt{L})}\!$};
     \draw[stateTransition,opacity=1] (zO) -- (y);
     
      \node (xt)[circle,inner sep=0]   at (0, 0.45) {\raisebox{3pt}{$\begin{array}{c}f(\bm{x})\\f(\bm{\theta})\end{array}$}};
    \node (z1t)[circle,inner sep=0] at (1, 0.45) {$f(\bm{\theta},\!\bm{z}^{\!(1)}\!)$};
        \node (zit) at (2, 0.45) {$\cdots$};
    \node (zLt)[circle,inner sep=0] at (3, 0.45) {$f(\bm{\theta},\!\bm{z}^{\!(\mathtt{L})}\!)$};
    \node (zOt)[circle,inner sep=0] at (4,0.45) {$f(\bm{\theta},\!\bm{z}^{\!(\mathtt{O})}\!)$};
    \node (yt)[circle,inner sep=0] at (4.85,0.45) {$f(\bm{\theta},\!\bm{y})$};
      \draw[Forward,opacity=1] (x) to [out=40,in=140] (z1);
      \draw[Forward,opacity=1] (z1) to [out=40,in=140] (zi);
      \draw[Forward,opacity=1] (zi) to [out=40,in=140] (zL);
      \draw[Forward,opacity=1] (zL) to [out=40,in=140] (zO);
      \draw[Forward,opacity=1] (zO) to [out=40,in=140] (y);
      
      \node[circle,inner sep=0] (xb) at (0.5,-0.45) {\raisebox{3pt}{$f(\bm{\theta}|\bm{y})$}};
     \node[circle,inner sep=0] (yb) at (4.85,-0.45) {\raisebox{1pt}{$f(\bm{y})$}};
        
      \draw[Inference,opacity=1] (yb) to [out=180,in=-27] (xp);
      \draw[Inference,opacity=1] (yb) to [out=180,in=-27] (z1p);
      \draw[Inference,opacity=1] (yb) to [out=180,in=-35] (zip);
      \draw[Inference,opacity=1] (yb) to [out=180,in=-60] (zLp);
      

\end{tikzpicture}}
\!\!\!\!\!\!\!\!\!\!\!\!\subfloat[][Tractable recursive layer-wise inference (Eq. \ref{EQ:output_inf}--\ref{EQ:RTS_inference_z})]{\tikzset{dist1/.style={path picture= {
    \begin{scope}[x=1pt,y=10pt]
      \draw plot[domain=-6:6] (\x,{1/(1 + exp(-\x))-0.5});
    \end{scope}
    }
  }
}
\tikzset{dist2/.style={path picture= {
    \begin{scope}[x=1pt,y=10pt]
      \draw plot[domain=-6:6] (\x,{\x/10});
    \end{scope}
    }
  }
}
\tikzstyle{input}=[draw,fill=red!10,circle,minimum size=15pt,inner sep=0pt]
\tikzstyle{hidden}=[draw,fill=green!20,circle,minimum size=15pt,inner sep=0pt]
\tikzstyle{output}=[draw,fill=blue!20,circle,minimum size=15pt,inner sep=0pt]
\tikzstyle{bias}=[draw=none,circle,minimum size=15pt,inner sep=0pt]

\tikzstyle{stateTransition}=[->,line width=0.25pt,draw=black!50]
\tikzstyle{Forward}=[->,line width=0.25pt,draw=magenta!75]
\tikzstyle{Inference}=[->,line width=0.25pt,draw=cyan!75]

\begin{tikzpicture}[scale=1.45]\scriptsize
    \node (x)[input]   at (0, 0) {$\bm{x}$};
    \node (z1)[hidden] at (1, 0) {$\,\bm{z}^{\!(1)}$};
    \node (zi) at (2, 0) {$\cdots$};
    \node (zL)[hidden] at (3, 0) {$\,\bm{z}^{\!(\mathtt{L})}$};
    \node (zO)[hidden] at (4,0) {$\,\bm{z}^{\!(\mathtt{O})}$};
    \node (y)[output] at (4.85,0) {$\bm{y}$};  
    
    \node (xp)[minimum size=7pt]   at (0.5, 0) {};
    \node (z1p)[minimum size=7pt] at (1.5, 0) {};
    \node (zip)[minimum size=7pt] at (2.5, 0) {};
    \node (zLp)[minimum size=7pt] at (3.5, 0) {};

    \draw[stateTransition,opacity=1] (x) -- (z1) node [midway, rotate=0,fill=white,opacity=0.8] {\scriptsize$\!\bm{\theta}^{(0)}\!$};
    \draw[stateTransition,opacity=1] (z1) -- (zi) node [midway, rotate=0,fill=white,opacity=0.8] {\scriptsize$\!\bm{\theta}^{(1)}\!$};
          \draw[stateTransition,opacity=1] (zi) -- (zL) node [midway, rotate=0,fill=white,opacity=0.8] {\scriptsize$\!\!\bm{\theta}^{(\mathtt{L}\text{-}1)}\!\!$};
    \draw[stateTransition,opacity=1] (zL) -- (zO) node [midway, rotate=0,fill=white,opacity=0.8] {\scriptsize$\!\bm{\theta}^{(\mathtt{L})}\!$};
     \draw[stateTransition,opacity=1] (zO) -- (y);
     
      \node (xt)[circle,inner sep=0]   at (0, 0.5) {\raisebox{3pt}{$\begin{array}{c}f(\bm{x})\\f(\bm{\theta})\end{array}$}};
        \node (z1t)[circle,inner sep=0] at (1, 0.5) {$\begin{array}{c}f(\bm{x},\!\bm{z}^{\!(1)}\!)\\f(\bm{\theta}^{(0)}\!\!,\!\bm{z}^{\!(1)}\!)\end{array}$};
        \node (zit) at (2, 0.45) {$\cdots$};
    \node (zLt)[circle,inner sep=0] at (3, 0.5) {$\begin{array}{c}f(\bm{z}^{\!(\mathtt{L}\text{-}1)}\!\!,\!\bm{z}^{\!(\mathtt{L})}\!)\\f(\bm{\theta}^{(\mathtt{L}\text{-}1)}\!\!,\!\bm{z}^{\!(\mathtt{L})}\!)\end{array}$};
    \node (zOt)[circle,inner sep=0] at (4,0.5) {$\begin{array}{c}f(\bm{z}^{\!(\mathtt{L})}\!\!,\!\bm{z}^{\!(\mathtt{O})}\!)\\f(\bm{\theta}^{(\mathtt{L})}\!\!,\!\bm{z}^{\!(\mathtt{O})}\!)\end{array}$};
    \node (yt)[circle,inner sep=0] at (4.85,0.5) {$f(\bm{z}^{\!(\mathtt{O})}\!\!,\!\bm{y})$};
      \draw[Forward,opacity=1] (x) to [out=40,in=140] (z1);
      \draw[Forward,opacity=1] (z1) to [out=40,in=140] (zi);
      \draw[Forward,opacity=1] (zi) to [out=40,in=140] (zL);
      \draw[Forward,opacity=1] (zL) to [out=40,in=140] (zO);
      \draw[Forward,opacity=1] (zO) to [out=40,in=140] (y);
      
      \node[circle,inner sep=0] (xb) at (0,-0.35) {$f(\bm{x}|\bm{y})$};
            \node[circle,inner sep=0] (z1b) at (1,-0.35) {$f(\bm{z}^{\!(\mathtt{1})}\!|\bm{y})$}; 
     \node (zib) at (2, -0.45) {$\cdots$};
      \node[circle,inner sep=0] (zLb) at (3,-0.35) {$f(\bm{z}^{\!(\mathtt{L})}\!|\bm{y})$}; 
      \node[circle,inner sep=0] (zOb) at (4,-0.35) {$f(\bm{z}^{\!(\mathtt{O})}\!|\bm{y})$}; 
     \node[circle,inner sep=0] (yb) at (4.85,-0.45) {\raisebox{1pt}{$f(\bm{y})$}};
     
      \node[circle,inner sep=0] (xb) at (0.5,-0.6) {$f(\bm{\theta}^{(0)}\!|\bm{y})$};
     \node[circle,inner sep=0] (z1b) at (1.5,-0.6) {$f(\bm{\theta}^{(1)}\!|\bm{y})$}; 
          \node[circle,inner sep=0] (z1b) at (2.5,-0.6) {$f(\bm{\theta}^{(\mathtt{L\text{-}1})}\!|\bm{y})$}; 
           \node[circle,inner sep=0] (zLpb) at (3.5,-0.6) {$f(\bm{\theta}^{(\mathtt{L})}\!|\bm{y})$}; 
               
      \draw[Inference,opacity=1] (z1) to [out=-140,in=-40] (x);
      \draw[Inference,opacity=1] (zi) to [out=-140,in=-40] (z1);
      \draw[Inference,opacity=1] (zL) to [out=-140,in=-40] (zi);
      \draw[Inference,opacity=1] (zO) to [out=-140,in=-40] (zL);
      \draw[Inference,opacity=1] (y) to [out=-140,in=-40] (zO);
      
      \draw[Inference,opacity=1] (z1) to [out=-140,in=-60] (xp);
      \draw[Inference,opacity=1] (zi) to [out=-140,in=-60] (z1p);
      \draw[Inference,opacity=1] (zL) to [out=-140,in=-60] (zip);
      \draw[Inference,opacity=1] (zO) to [out=-140,in=-60] (zLp);

\end{tikzpicture}}
\caption{Representation of the \emph{forward} propagation of uncertainty ({\color{magenta!75}magenta arrows}) and \emph{inference} procedures ({\color{cyan!75}cyan arrows}).}
\label{FIG:FNN_inference_long}
\end{figure}
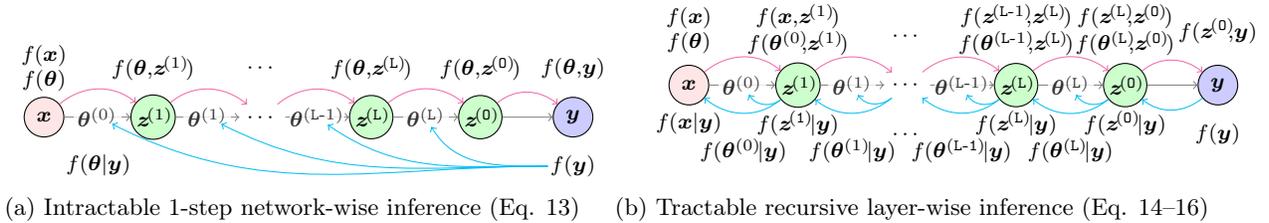}

The solution we propose to overcome these challenges is twofold: (1)  employ a diagonal covariance structure for both the parameters $\bm{\theta}$, and hidden units $\bm{Z}^{(j)}$, and (2) use the inherent conditional independence of hidden units between layers, that is, $\bm{Z}^{(j-1)}\independent\bm{Z}^{(j+1)}|\bm{z}^{(j)}$, in order to perform recursive layer-wise Gaussian inference. As depicted in Figures \ref{FIG:FNN_inference_long} \& \ref{FIG:FNN}, in the feedforward process, there is no direct information path between hidden layers $j-1$ and $j+1$ other than through the hidden layer $j$. Therefore, the knowledge of hidden units at layer $j$ blocks the information from the layer $j-1$, so that layers $j+1$ and  $j-1$ are conditionally independent under the assumptions that parameters $\bm{\theta}$ are independent between layers. As depicted in Figure \ref{FIG:FNN_inference_long}b by the rightmost blue arrow from $\bm{y}$ to $\bm{z}^{(\mathtt{O})}$, the first step consists in inferring the posterior mean vector and diagonal covariance for the output layer following 
\begin{equation}
\begin{array}{rcl}
f(\bm{z}^{(\mathtt{O})}|\bm{y})&=&\mathcal{N}(\bm{z}^{(\mathtt{O})};\bm{\mu}_{\bm{Z}^{(\mathtt{O})}|\bm{y}},\bm{\Sigma}_{\bm{Z}^{(\mathtt{O})}|\bm{y}})\\[6pt]
\bm{\mu}_{\bm{Z}^{(\mathtt{O})}|\bm{y}}&=&\bm{\mu}_{\bm{Z}^{(\mathtt{O})}}+\mathbf{\Sigma}_{\bm{Y\!Z}^{(\mathtt{O})}}^{\intercal}\mathbf{\Sigma}_{\bm{Y}}^{-1}(\bm{y}-\bm{\mu}_{\bm{Y}})\\[4pt]
\mathbf{\Sigma}_{\bm{Z}^{(\mathtt{O})}|\bm{y}}&=&\mathbf{\Sigma}_{\bm{Z}^{(\mathtt{O})}}-\mathbf{\Sigma}_{\bm{Y\!Z}^{(\mathtt{O})}}^{\intercal}\mathbf{\Sigma}_{\bm{Y}}^{-1}\mathbf{\Sigma}_{\bm{Y\!Z}^{(\mathtt{O})}}.\end{array}
\label{EQ:output_inf}
\end{equation}
This step is analogous to performing message passing on the graph presented in Figure \ref{FIG:FNN}, where an observation is only available for the output node at the right end. Note that for classification problems, because of the hierarchical formulation described in Appendix \ref{A:binary_tree}, even if there are $\mathtt{Y}$ classes, only $\mathtt{H}=\lceil\log_2(\mathtt{Y})\rceil$ hidden units from the output layer are updated for each observation.

The layer-wise Gaussian inference for hidden units and parameters is again analogous to performing message passing on the graph in Figure \ref{FIG:FNN}, but this time, from the output layer to input. The backward passing of information depicted by blue arrows in Figure \ref{FIG:FNN_inference_long}b, is done using the Rauch-Tung-Striebel recursive procedure that was developed in the context of state-space models \citep{rauch1965maximum}. For the RTS procedure, we define the short-hand notation $\{\bm{\theta}^{\text{+}},\bm{Z}^{\text{+}}\}\equiv\{\bm{\theta}^{(j+1)},\bm{Z}^{(j+1)}\}$ and $\{\bm{\theta},\bm{Z}\}\equiv\{\bm{\theta}^{(j)},\bm{Z}^{(j)}\}$ so that
\begin{equation}
\begin{array}{rcl}
f(\bm{z}|\bm{y})&=&\mathcal{N}(\bm{z};\bm{\mu}_{\bm{Z}|\bm{y}},\bm{\Sigma}_{\bm{Z}|\bm{y}})\\[4pt]
\bm{\mu}_{\bm{Z}|\bm{y}}&=&\bm{\mu}_{\bm{Z}}+\mathbf{J}_{\bm{Z}}\left(\bm{\mu}_{\bm{Z}^{\text{+}}|\bm{y}}-\bm{\mu}_{\bm{Z}^{\text{+}}}\right)\\[4pt]
\bm{\Sigma}_{\bm{Z}|\bm{y}}&=&\bm{\Sigma}_{\bm{Z}}+\mathbf{J}_{\bm{Z}}\left(\bm{\Sigma}_{\bm{Z}^{\text{+}}|\bm{y}}-\bm{\Sigma}_{\bm{Z}^{\text{+}}}\right)\mathbf{J}_{\bm{Z}}^{\intercal}\\[4pt]
\mathbf{J}_{\bm{Z}}&=&\bm{\Sigma}_{\bm{Z}\!\bm{Z}^{\text{+}}}\bm{\Sigma}_{\bm{Z}^{\text{+}}}^{-1},\end{array}\label{EQ:RTS_inference_y}\end{equation}
\begin{equation}
\begin{array}{rcl}
f(\bm{\theta}|\bm{y})&=&\mathcal{N}(\bm{\theta};\bm{\mu}_{\bm{\theta}|\bm{y}},\bm{\Sigma}_{\bm{\theta}|\bm{y}})\\[4pt]
\bm{\mu}_{\bm{\theta}|\bm{y}}&=&\bm{\mu}_{\bm{\theta}}+\mathbf{J}_{\bm{\theta}}\left(\bm{\mu}_{\bm{Z}^{\text{+}}|\bm{y}}-\bm{\mu}_{\bm{Z}^{\text{+}}}\right)\\[4pt]
\bm{\Sigma}_{\bm{\theta}|\bm{y}}&=&\bm{\Sigma}_{\bm{\theta}}+\mathbf{J}_{\bm{\theta}}\left(\bm{\Sigma}_{\bm{Z}^{\text{+}}|\bm{y}}-\bm{\Sigma}_{\bm{Z}^{\text{+}}}\right)\mathbf{J}_{\bm{\theta}}^{\intercal}\\[4pt]
\mathbf{J}_{\bm{\theta}}&=&\bm{\Sigma}_{\bm{\theta}\bm{Z}^{\text{+}}}\bm{\Sigma}_{\bm{Z}^{\text{+}}}^{-1}.
\end{array}
\label{EQ:RTS_inference_z}
\end{equation}
The key aspect of this layer-wise approach is that in the forward step depicted in Figure \ref{FIG:FNN_inference_long}b by magenta arrows, we only need to store the layer-wise mean vectors $\{\bm{\mu}_{\bm{\theta}},\bm{\mu}_{\bm{Z}}\}$ and covariances $\{\bm{\Sigma}_{\bm{\theta}},\bm{\Sigma}_{\bm{Z}},\bm{\Sigma}_{\bm{\theta}\bm{Z}^{\text{+}}},\bm{\Sigma}_{\bm{Z}\!\bm{Z}^{\text{+}}}\}$, where in addition to the relations already given in \S\ref{SS:LAFFNN}, the cross-covariance matrix for two successive layers is $$\bm{\Sigma}_{\bm{Z}\!\bm{Z}^{\text{+}}}=\mathbf{F}_{\bm{wa}}^{(j)}\text{cov}(\bm{W\!\!A}^{(j)}\!,\!\bm{Z}^{(j)})+\mathbf{F}_{\bm{b}}^{(j)}\text{cov}(\bm{B}^{(j)}\!,\!\bm{Z}^{(j)}).$$
With a diagonal covariance structure for both $\bm{Z}$ and $\bm{\theta}$, the covariance matrices defining each layer contain at most $\mathtt{A}^2+\mathtt{A}$ non-zero terms, i.e., the number of weights ($\mathtt{A}^2$) and biases ($\mathtt{A}$) per layer; Because of the diagonal structures of covariance matrices, equations \ref{EQ:output_inf}--\ref{EQ:RTS_inference_z} have a computational complexity $\mathcal{O}(\mathtt{A}^2)$, which scales linearly with the number of hidden layers $\mathtt{L}$.

The inference of parameters $\bm{\theta}$ is done recursively so that after having seen either a single observation or a batch of them, the posterior becomes the prior for the next observations. In order to learn from batches consisting of multiple observations, one can perform the forward propagation of uncertainty for several observations, all sharing the same hyperparameters ${\bm{\eta}}=\{\bm{\mu}_{\bm{\theta}},\mathbf{\Sigma}_{\bm{\theta}}\}$, and then do a single update for all of them. As it is the case for deterministic neural networks, the batch procedure allows leveraging parallel computing.

 One aspect that needs to be considered is that the finals result of the inference will depend on the initialization of hyperparameters $\bm{\eta}$,  as well as on data ordering. It happens because neural networks are inherently non-identifiable so that the true posterior for their parameters is multimodal. For example, we can imagine how any permutation of hidden units in a same layer that would not change the results but would lead to a different posterior. Therefore, TAGI depends upon hyperparameter initialization and data ordering because it approximates a multimodal posterior with an unimodal multivariate Gaussian.

\subsection{Hyper-parameter estimation}
There are typically tens of thousands, if not millions, of parameters in $\bm{\theta}$, for which we typically have little or no prior information for defining the hyper-parameters ${\bm{\eta}}^{(\texttt{0})}=\{\bm{\mu}_{\bm{\theta}}^{(\texttt{0})},\mathbf{\Sigma}_{\bm{\theta}}^{(\texttt{0})}\}$. In the case where we have small datasets, the weakly informative prior combined with limited data will lead to a weakly informative posterior. One solution to go around this difficulty while avoiding overfitting is to learn the model parameters over multiple epochs, $\mathtt{E}>1$, using a training $\mathcal{D}_{\mathtt{T}}$ and validation set $\mathcal{D}_{\mathtt{V}}$. Here, we propose to employ the posterior's hyper-parameter values at the $i^{th}$ iteration ${\bm{\eta}}^{(i)}=\{\bm{\mu}_{\bm{\theta}|\mathcal{D}_{\mathtt{T}}}^{(i)},
\mathbf{\Sigma}_{\bm{\theta}|\mathcal{D}_{\mathtt{T}}}^{(i)}\}$ and use them as the prior's hyper-parameters at the next iteration $i+1$. This recursive procedure is stopped when the marginal likelihood $f(\mathcal{D}_{y,\mathtt{V}}|\mathcal{D}_{x,\mathtt{V}},{\bm{\eta}}^{(i)})\,=\,\mathcal{N}\big(\mathcal{D}_{\mathtt{V}};\bm{\mu}_{\bm{y}_{\mathtt{v}}|\mathcal{D}_{\mathtt{T}}}^{(i)},
\mathbf{\Sigma}_{\bm{y}_{\mathtt{v}}|\mathcal{D}_{\mathtt{T}}}^{(i)}\big)$ for the validation set $\mathcal{D}_{\mathtt{V}}$, has reached its maximal value. This procedure is analogous to the empirical Bayes approach \citep{efron2012large} where the prior knowledge's hyper-parameters are learnt through the maximization problem \begin{equation}\hat{\bm{\eta}}=\underset{\bm{\eta}}{\arg\,\max}~\int f(\mathcal{D}_{y,\mathtt{V}}|\mathcal{D}_{x,\mathtt{V}},\bm{\theta})\cdot f(\bm{\theta}|\bm{\eta})d\bm{\theta}.\label{EQ:empirical_bayes}\end{equation}
Note that unlike in Equation \ref{EQ:empirical_bayes} where the maximization is explicit, in our case the maximization is implicitly performed by updating over multiple epochs. Even if the recursive approach proposed is not guaranteed to lead to the Type-2 maximum likelihood estimate solution sought by empirical Bayes, it is much more efficient than having to perform the explicit maximization for the expected value and variance parameters. 

\section{Related work}\label{S:related_work}
Many researchers have already proposed approximate inference methods for \emph{Bayesian Neural Networks} (BNN). Early on, it was proposed to employ \emph{message passing} methods (\citeauthor{murphy2012machine}, 2012, \S18.3) such as the \emph{Extended} \citep{NIPS1988_101, puskorius1991decoupled}, \emph{Unscented Kalman Filter} \citep{wan2000unscented}, and the \emph{RTS smoother} \citep{haykin2004kalman} to leverage Gaussian inference in order to approximate the posterior. The main issue with these approaches is related to their computational complexity which is proportional to the square of the number of weights \citep{wan2000unscented}. More recently, this Gaussian inference framework was extended with the \emph{Cubature filter} \citep{arasaratnam2008nonlinear}, which, like the extended and unscented methods, is limited by its computational complexity. The main factor limiting the computational efficiency of these approaches is the usage of full covariance matrices for the network parameters. \cite{plumer1995training} has proposed to employ either fully-decoupled or layer-decoupled variants, but the loss in performance prevents their recommendation as a default choice \citep{haykin2004kalman}.  

In another early approach, \citeauthor{mackay1992practical} (\citeyear{mackay1992practical}) employed the \emph{Laplace approximation}  to describe the posterior covariance of parameters. The author experimented with neural networks having a small number of hidden units per layer (i.e., $5$--$20$) and noted that ``\emph{diagonal approximation are no good because of the strong posterior correlation}''. Later, \citeauthor{neal1995Bayesian}  (\citeyear{neal1995Bayesian}) explored the potential of BNN using the Hamiltonian Monte Carlo (HMC) method. Despite its lack of practical scalability in the context of large neural networks applied to image classification tasks, natural language processing, or reinforcement learning, HMC is seen as a classic reference method for Bayesian inference \citep{gelman2014bayesian}. For instance, \cite{NEURIPS2020_2dfe1946} have employed HMC and other structured-covariance methods to demonstrate empirically that there is no significant difference in performance when using a diagonal or a  full covariance structure for the weight parameters of deep neural networks.

In parallel, several researchers have applied \emph{variational inference} for estimating the posterior distribution of neural networks'\! parameters \citep{hinton1993keeping,barber1998ensemble}. The development of moment matching and variational approaches for BNN is still nowadays an active research area \citep{NIPS2015_bc731692,hernandez2015probabilistic,blundell2015weight,louizos2016structured,Osawa2019PracticalDL,DBLP:confWuNMTHG19}. Recently, the technique of using Dropout as a Bayesian approximation has received a lot of attention in the community \citep{gal2016dropout}. All the recent methods that are either based on moment matching, variational approaches, or dropout, share a common aspect; the inference of parameters is still treated as an optimization problem relying on gradient backpropagation. Although some approaches such as the one by Wu et al. (\citeyear{DBLP:confWuNMTHG19}) have exploited approximate analytically tractable mean and variance propagation under Gaussian assumptions, it still relies on variational inference, as none of the approaches currently available allow for analytically tractable inference in neural networks, while reaching a competitive level of performance and efficiency.

\section{Experiments}\label{S:experiments}
In this section, we perform experiments using the TAGI method for a 1D toy problem, for a set of benchmark regression problems, and for the MNIST classification dataset.
\subsection{1D toy problem}\label{SS:toy_problem}
We apply TAGI to the 1D regression problem $y=x^{3}+v$, such that $V\sim\mathcal{N}(0,9)$, as taken from \citep{hernandez2015probabilistic}, using an AG-FNN having a single hidden layer with 100 units, and ReLU activation functions. The objective of this case study is to showcase how TAGI can be applied on small datasets ($\mathtt{D}=20$ points), and to compare the results obtained by considering either diagonal or full covariance matrices. In this example, the inference is performed using one observation at a time, where both the covariates $x$ and observations $y$ were normalized in the range $[-1,1]$. The optimal number of epochs is identified from a validation set $\mathcal{D}_{\mathtt{V}}$ consisting of 20 additional points. The prior covariance for bias is initialized to $\bm{\Sigma}_{\bm{B}}^{\texttt{0}}=0.01\cdot\mathbf{I}$, and for weights $\bm{\Sigma}_{\bm{W}}^{\texttt{0}}$, by using the Xaviers's approach \citep{glorot2010understanding}. The prior mean vector is randomly sampled from $\bm{\mu}_{\bm{\theta}}^{\texttt{0}}\sim\mathcal{N}(\mathbf{0},\bm{\Sigma}_{\bm{\theta}}^{\texttt{0}})$ in order to break the initial symmetry in the network, as starting with zero expected values for weights adversely affects learning.

\begin{figure}[h!]
\centering
\subfloat[][TAGI, $\mathtt{E}\!=\!0$]{{\includegraphics[height=27.5mm]{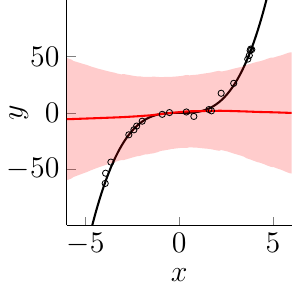}}}~\subfloat[][TAGI, $\mathtt{E}\!=\!1$]{{\includegraphics[height=27.5mm]{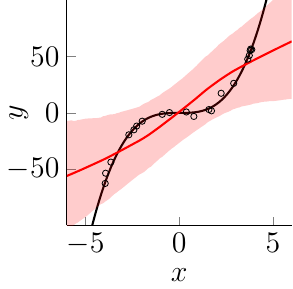}}}~\subfloat[][TAGI, $\mathtt{E}\!=\!50$]{{\includegraphics[height=27.5mm]{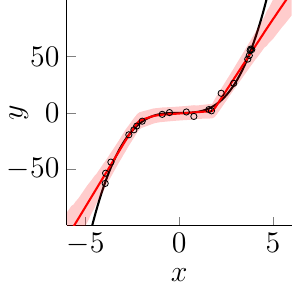}}}\\[-6pt]\subfloat[][TAGI, $\mathtt{E}\!=\!31$]{{\includegraphics[height=28mm]{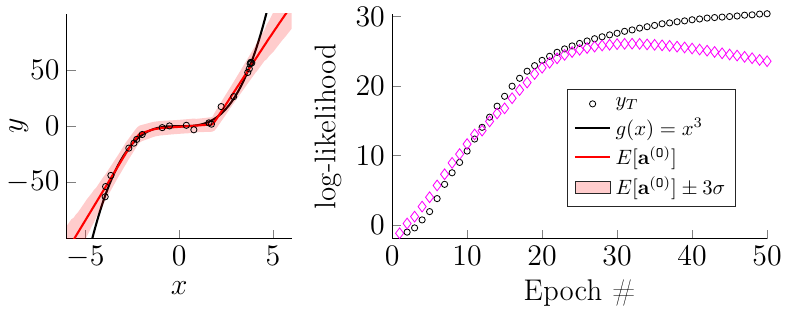}}}~\subfloat[][HMC]{{\includegraphics[height=27.5mm]{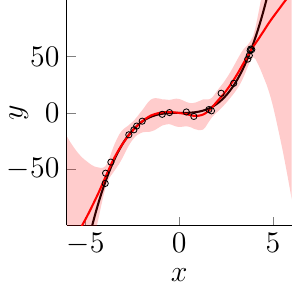}}}
\caption{Application of AG-FNN with diagonal covariances to a toy problem where (a--c) describe the evolution of the predictive distribution with respect to the number of epochs $\mathtt{E}$. In (d), we compare the training and validation log-likelihood in order to identify the optimal number of epochs, i.e., $\mathtt{E}=31$. In (e) we compare with the results obtained using HMC.} 
\label{fig:1D_reg}
\end{figure}

Figure \ref{fig:1D_reg} compares the true function employed to generate the data, with the AG-FNN (with diagonal covariances) predictions described by their expected values and $\pm3\sigma$ confidence regions. We can see in (a) that the prior predictive obtained before updating with observations ($\mathtt{E}=0$) is weakly informative, and that the posterior predictive obtained after the first epoch (b, $\mathtt{E}=1$) is still a poor approximation of the true function. The log-likelihood reported in (d) for the validation set allows identifying that the optimal number of epochs is here $\mathtt{E}=31$. The log-likelihood values  reported in Figure \ref{fig:1D_reg}d confirm that employing a training set to identify the number of epochs would not be able to prevent overfitting as depicted in (c). We can see in (e) the same results processed with Hamiltonian Monte-Carlo (HMC) using the implementation by \cite{cobb2019introducing}. One aspect we observe is that TAGI is not able to correctly extrapolate the widening confidence region outside the training data, as correctly displayed by HMC. The main reason behind this behaviour is that TAGI only provide an unimodal posterior; In the case of HMC, the multiple posterior modes all coincide to similar predictions when constrained by data but not during extrapolation.

\begin{figure}[b!]\subfloat[][$\mathtt{L}\!=\!1$, $\mathtt{A}\!=\!5$]{\includegraphics[width=0.5 \columnwidth]{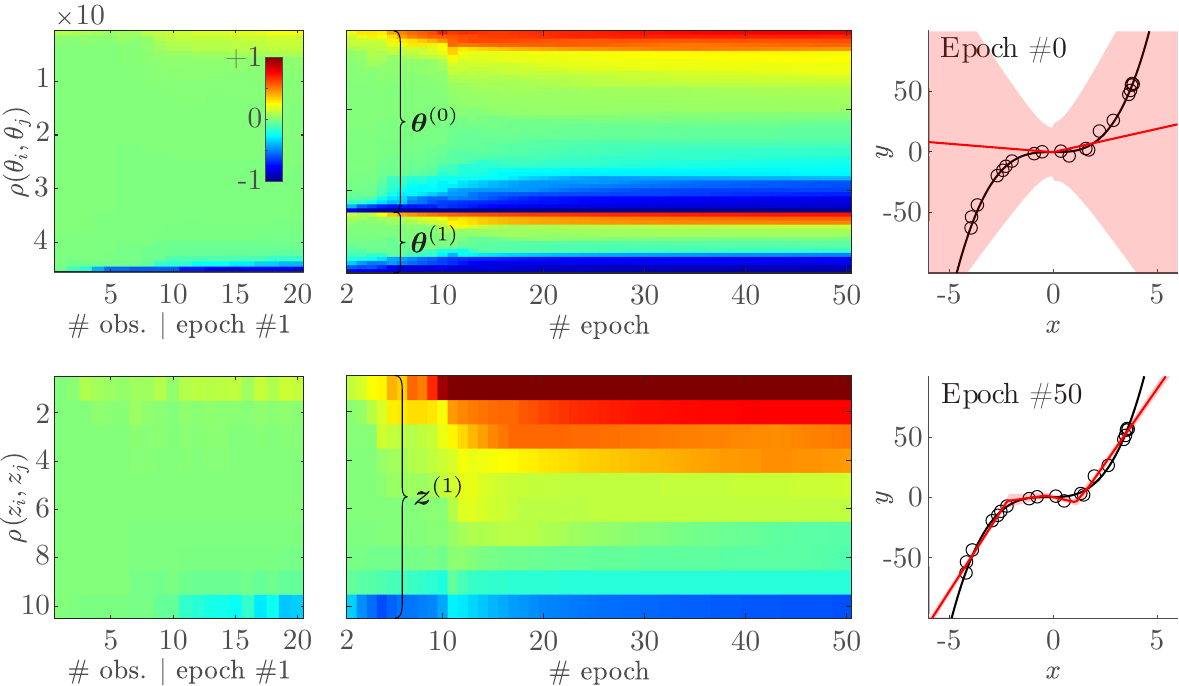}}~
\subfloat[][$\mathtt{L}\!=\!2$, $\mathtt{A}\!=\!5$]{\includegraphics[width=0.5 \columnwidth]{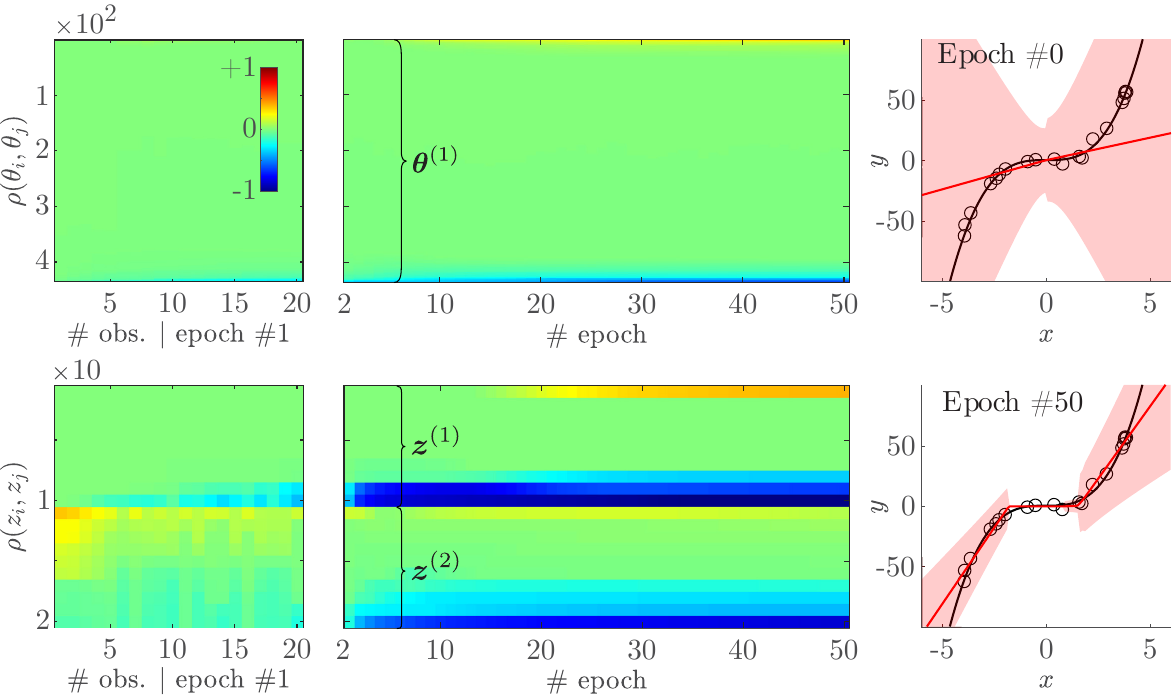}}\\[-6pt]
\!\!\!\!\subfloat[][$\mathtt{L}\!=\!1$, $\mathtt{A}\!=\!50$]{\includegraphics[width=0.5 \columnwidth]{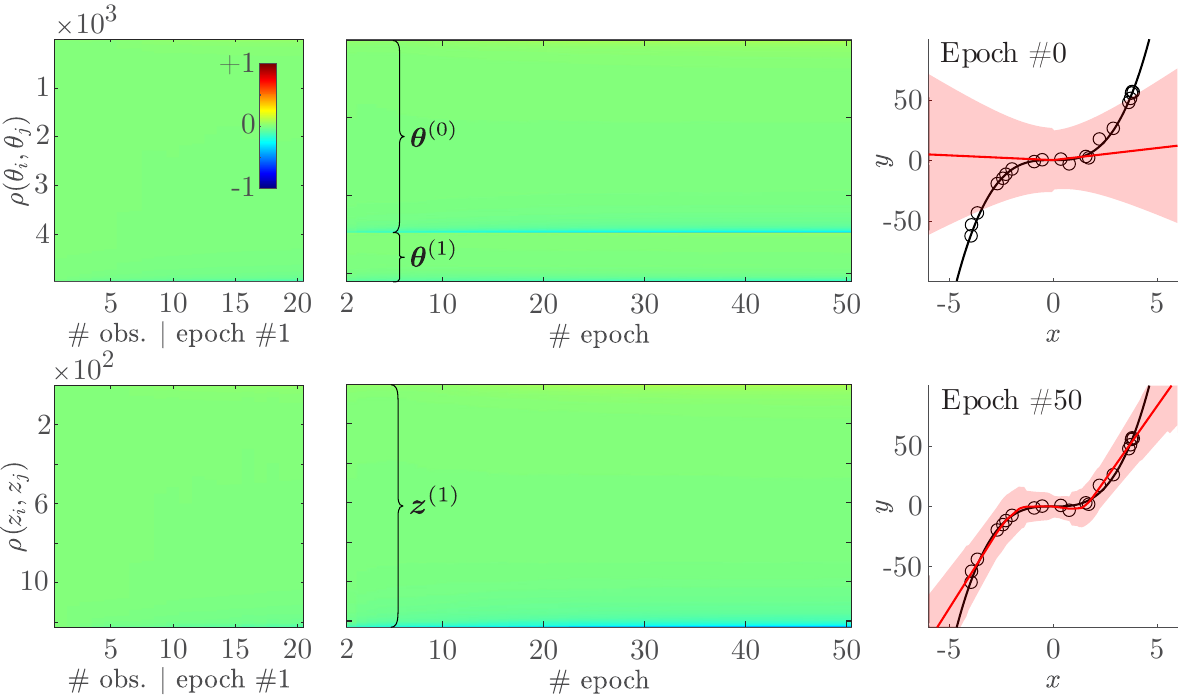}}~
\subfloat[][$\mathtt{L}\!=\!2$, $\mathtt{A}\!=\!50$]{\includegraphics[width=0.5 \columnwidth]{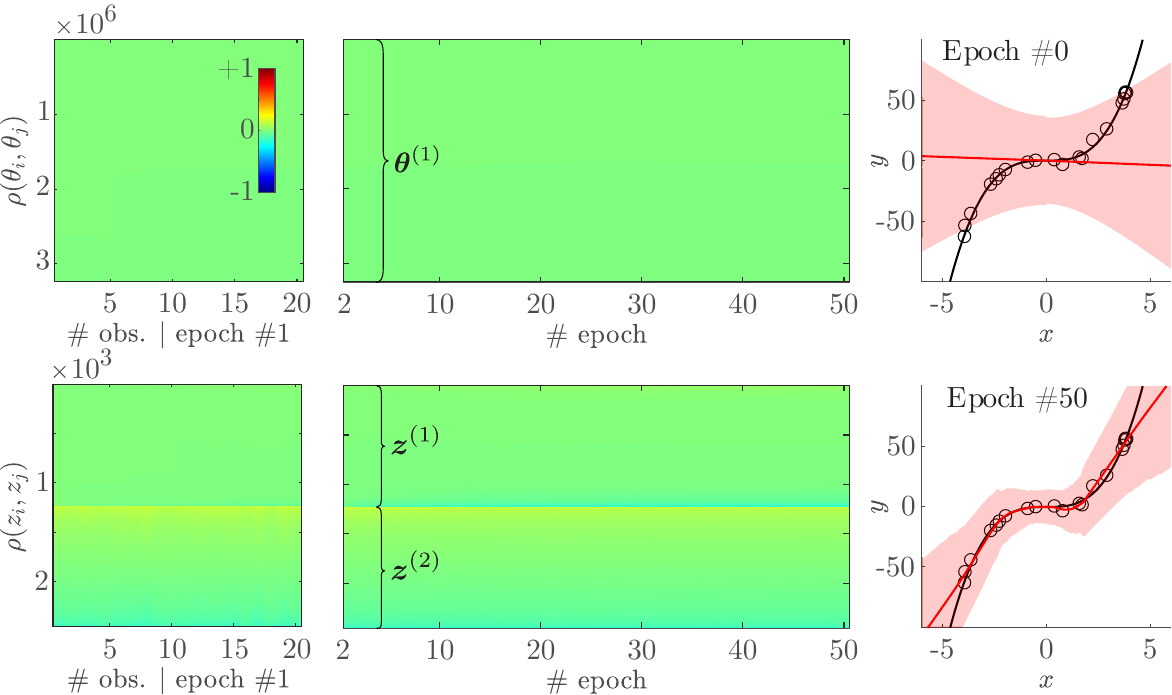}}
\caption{Representation of the sorted correlation coefficients extracted from upper-triangular posterior covariance matrices for the parameters $\bm{\theta}^{(j)}$ and hidden units $\bm{z}^{(j)}$. The left-most graphs present the correlations for each of the 20 observations within the first epoch, and the center graphs present the correlations at the end of each epoch.}
\label{FIG:full_covariance}
\end{figure}
In a second experiment, we now apply TAGI to the same dataset while considering the full covariance matrices for $\bm{\theta}^{(j)}$ and $\bm{z}^{(j)}$. We study networks with either $\mathtt{L}\!=\!1$ or $2$ hidden layers, each comprising either $\mathtt{A}\!=\!5$ or $50$ hidden units. Figure \ref{FIG:full_covariance} displays colour maps where each pixel represents the layer-wise sorted correlation coefficients extracted from the upper-triangular portion of the posterior covariance matrices, for either the parameters $\bm{\theta}^{(j)}$ (top), or the hidden units $\bm{z}^{(j)}$ (bottom). Note that in the cases (b) and (d), results are only presented for the fully connected layer $\bm{\theta}^{(1)}$ because of the disproportionate number of weights in it in comparison with the input and output layers. The results displayed on the left colour map from each subfigure corresponds to each observation from the first epoch, and the right colour maps are for the last observation from subsequent epochs. 

\begin{figure}[htbp]\centering
\includegraphics[width=0.45 \columnwidth]{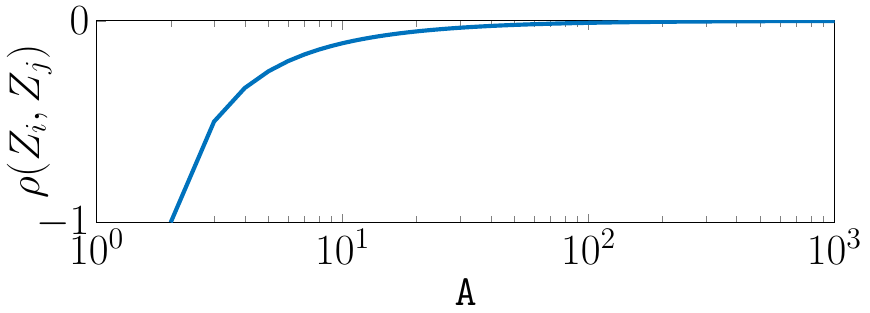}\\[-6pt]
\caption{Evolution of the posterior correlation $\rho(Z_i,Z_j)$ as a function of the number of variables $\mathtt{A}$ for an observation model $y=\sum_{i=1}^\mathtt{A}z_i$, where all  $Z_i\sim\mathcal{N}(z,0,1)$ are independent.}
\label{FIG:Gaussian_conditional_rho}
\end{figure}

In Figure \ref{FIG:full_covariance}a,b for hidden layers of $\mathtt{A}=5$ units, we notice the presence of extensive large positive and negative correlations in the posteriors. In Figure \ref{FIG:full_covariance}c,d for hidden layers of $\mathtt{A}=50$ units, non-zero correlations, i.e., those with a colour other than green, are restricted to a small fraction of the covariance matrix. This can be explained by the theoretical results presented in Figure \ref{FIG:Gaussian_conditional_rho_prior}, and by further looking at the simplified context for the sum of several independent inputs $Z_i\sim\mathcal{N}(z;0,1)$ such that the observation model is $y=\sum_{i=1}^\mathtt{A}z_i$. If we observe a constant $y\in\mathbb{R}$, we can then exactly infer the Gaussian posterior for $\bm{Z}$, for which the correlations $\rho(Z_{i},Z_j),~\forall i\neq j$ are directly impacted by the number of variables $\mathtt{A}$ as depicted in Figure \ref{FIG:Gaussian_conditional_rho}. For small numbers of hidden units, i.e. $<10$, the posterior correlation is non-negligible, whereas, for large numbers, i.e. $>100$, it is almost zero. This confirms what was shown theoretically in Figure \ref{FIG:Gaussian_conditional_rho_prior}, that if the number of units per layer $\mathtt{A}$ is sufficient, the hidden units among a layer are independent, and thus considering only diagonal covariance matrices may lead to results that are comparable to those obtained while considering the full covariance structure. 

\begin{figure}[t!]\centering\vspace{-3mm}
\!\subfloat[][{$\mathtt{L}\!=\!1$,\,$\mathtt{A}\!=\!50$,\,$\bm{\Sigma_Z}\!\!:$\,full}]{\includegraphics[width=0.24 \columnwidth]{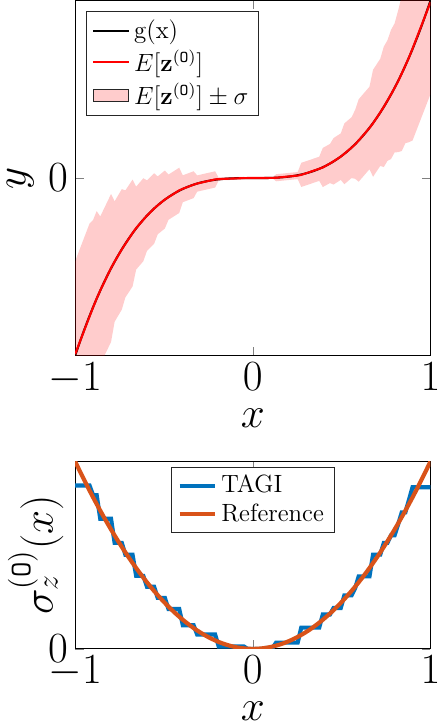}}~~
\subfloat[][$\mathtt{L}\!=\!2$,\,$\mathtt{A}\!=\!50$,\,$\bm{\Sigma_Z}\!\!:$\,full]{\includegraphics[width=0.24\columnwidth]{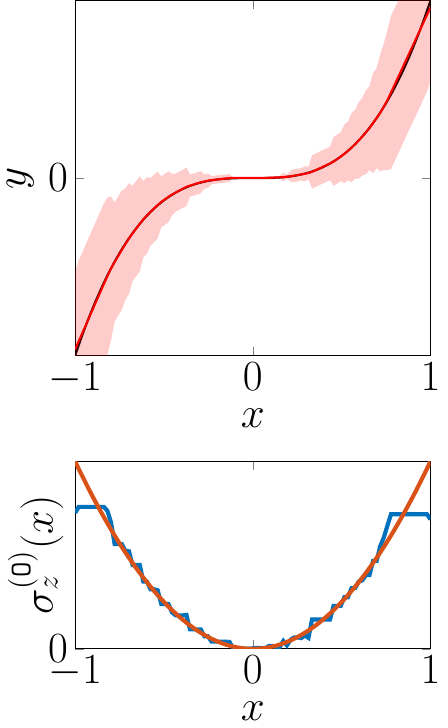}}~~
\subfloat[][$\mathtt{L}\!=\!1$,\,$\mathtt{A}\!=\!50$,\,$\bm{\Sigma_Z}\!\!:$\,diag]{\includegraphics[width=0.24 \columnwidth]{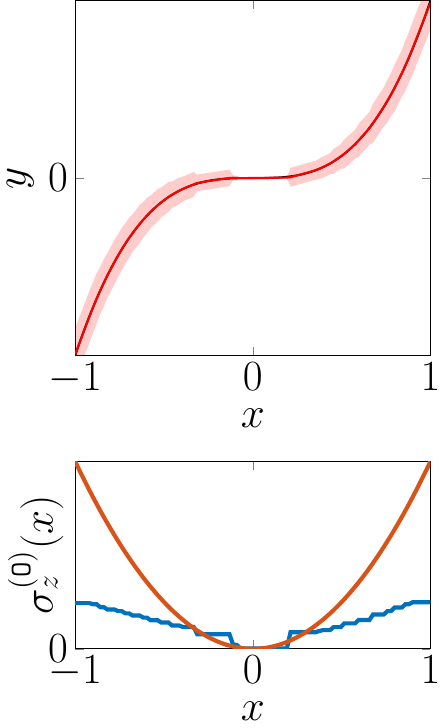}}~~
\subfloat[][$\mathtt{L}\!=\!2$,\,$\mathtt{A}\!=\!50$,\,$\bm{\Sigma_Z}\!\!:$\,diag]{\includegraphics[width=0.24\columnwidth]{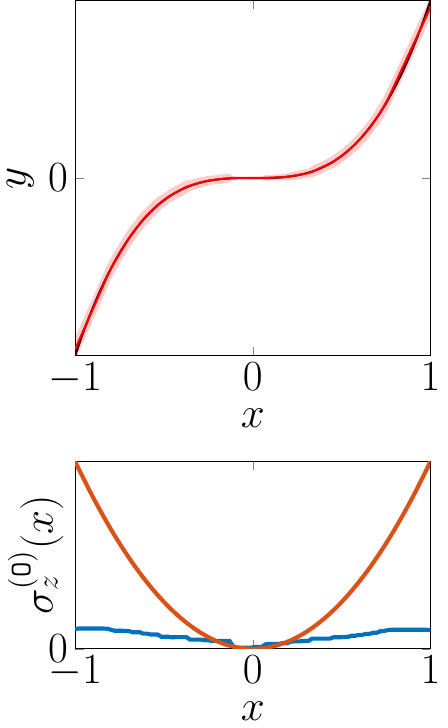}}
\caption{Comparison of the capacity to propagate uncertainty from the input to the output of a neural network while using either a full or a diagonal covariance $\bm{\Sigma_Z}$. The top graphs represent the neural network output with confidence interval where the bottom graphs compare the reference values with the empirical ones.}
\label{FIG:full_covariance_Sz}
\end{figure}

The third and last experiment exposes the limitations of using diagonal covariance matrices for hidden units, in the case where we care about the correct propagation of the input-layer uncertainty through a neural network. We use again the function $g(x)=x^3$, but this time in the range $x\in[-1,1]$, and with infinitely many exact observations so that $\bm{\Sigma_\theta}\to0$ and the parameters can be considered as deterministic. Once we have learnt the parameters exactly, we want to test the capacity to propagate the input-layer uncertainty for any value $x$ subject to an observation uncertainty such that $\tilde{X}\sim\mathcal{N}(x;0,\sigma_X^2)$, where $\sigma_X=\tfrac{2}{10}$. As depicted in Figure \ref{FIG:full_covariance_Sz}, because we use a ReLU activation function, the neural network's output is a piecewise-linear function and the reference conditional output standard deviation for such a locally linearized function can be computed using the $1^{st}$ order Taylor expansion so that 
\begin{equation}
\sigma_{Z}^{(\mathtt{O})}(x)=\sqrt{\text{var}[g(\tilde{X})|x]}=\left|\frac{dg(x)}{dx}\right|\cdot \sigma_X=\frac{6x^2}{10}.\label{EQ:sx}\end{equation}
The results in Figure \ref{FIG:full_covariance_Sz} are for both, full or diagonal hidden variable covariances $\bm{\Sigma_Z}$, and for one or two hidden layers. We can see that with a full $\bm{\Sigma_Z}$, the neural network's output uncertainty closely follow the reference values for $\sigma_{Z}^{(\mathtt{O})}(x)$. When using diagonal covariances for hidden layers, $\bm{\Sigma_Z}=\text{diag}(\bm{\sigma_Z})$, we are unable to retrieve the same reference values as defined in Equation \ref{EQ:sx}. Therefore, despite having attested in the previous experiments that TAGI can learn the parameters while using diagonal covariance matrices, it remains unable to correctly propagate uncertainty from the input layer to the output. For applications where propagating the input-layer uncertainty is critical, e.g. in partially observable reinforcement learning \citep{sutton2011reinforcement,pmlr-v80-jin18c}, one may choose to sacrifice the linear computational complexity of TAGI (see \S\ref{SS:bayesian_estimation}) by including full $\bm{\Sigma_Z}$ matrices for each layer during the forward propagation of uncertainty.  

\subsection{Benchmark regression problems}
The performance of TAGI is now compared with PBP \citep{hernandez2015probabilistic}, VMG \citep{louizos2016structured}, and MC-dropout \citep{gal2016dropout} for benchmark regression datasets. For the purpose of comparison with existing results taken from the literature, all the datasets are analyzed for a fixed number of epochs, that is $\mathtt{E}=40$. For all cases, the data is normalized, the activation function is a ReLU, and the batch size is $\mathtt{B}=10$; The prior covariance for biases is initialized to $\bm{\Sigma}_{\bm{B}}^{0}=0.01\cdot\mathbf{I}$, and for weights $\bm{\Sigma}_{\bm{W}}^{0}$, by using the Xaviers's approach \citep{glorot2010understanding}; The initial value for the observation error's standard deviation is set to $\sigma_{V}=1$, and this value is optimized using a 5-folds cross-validation setup. 

\begin{table*}[b]\scriptsize
\caption{Comparison TAGI's results with those available in the literature for PBP \citep{hernandez2015probabilistic}, VMG \citep{louizos2016structured}, and MC-dropout \citep{gal2016dropout} (Rank legend: {\bf\color{orange} first}, {\bf{\color{teal}second}}). The $\pm\sigma$ represent the standard deviation over the 20 test-folds.\\[0pt]}\centering
\scalebox{0.875}{\begin{tabular}{r|cccc|cccc}
\toprule
&\multicolumn{4}{c|}{Root mean square error (RMSE)}&\multicolumn{4}{c}{Average log-likelihood (LL)}\\
Datasets&PBP&VMG&MC-Dropout&{\bf TAGI}&PBP&VMG&MC-Dropout&{\bf TAGI}\\[2pt]
\cmidrule{1-9}
Boston 	&3.01$\pm$0.18&\bf\color{orange}2.70$\pm$0.13&\bf\color{teal}2.97$\pm$0.85&2.98$\pm$0.86 &\bf\color{teal}-2.56$\pm$0.12&\bf\color{orange}-2.46$\pm$0.09&\bf\color{orange}-2.46$\pm$0.25&-2.58$\pm$0.45\\
Concrete 	&5.67$\pm$0.09&\bf\color{orange}4.89$\pm$0.12&\bf\color{teal}5.23$\pm$0.53&5.72$\pm$0.52&-3.14$\pm$0.11&\bf\color{orange}-3.01$\pm$0.03&\bf\color{teal}-3.04$\pm$0.09&-3.17$\pm$0.09\\
Energy 	&1.80$\pm$0.05&\bf\color{orange}0.54$\pm$0.02&1.66$\pm$0.19&\bf\color{teal}1.46$\pm$0.22&-2.04$\pm$0.02&\bf\color{orange}-1.06$\pm$0.03&-1.99$\pm$0.09&\bf\color{teal}-1.81$\pm$0.14\\
Kin8nm 	&0.10$\pm$0.00&0.10$\pm$0.00&0.10$\pm$0.00&0.10$\pm$1E-3&0.90$\pm$0.01&\bf\color{orange}1.10$\pm$0.01&\bf\color{teal}0.95$\pm$0.03&0.88$\pm$0.04\\
Naval  &0.01$\pm$0.00&\bf\color{orange}0.00$\pm$0.00&0.01$\pm$0.00&0.01$\pm$6E-3&\bf\color{teal}3.73$\pm$0.01&2.46$\pm$0.12&\bf\color{orange}3.80$\pm$0.05&2.10$\pm$0.57\\
Power 	&4.12$\pm$0.03&\bf\color{teal}4.04$\pm$0.04&\bf\color{orange}4.02$\pm$0.18&4.12$\pm$0.16&-2.84$\pm$0.01&\bf\color{teal}-2.82$\pm$0.01&\bf\color{orange}-2.80$\pm$0.05&-2.83$\pm$0.04\\
Protein 			&4.73$\pm$0.01&\bf\color{orange}4.13$\pm$0.02&\bf\color{teal}4.36$\pm$0.04&4.70$\pm$0.02&-2.97$\pm$0.00&\bf\color{orange}-2.84$\pm$0.00&\bf\color{teal}-2.89$\pm$0.01&-2.97$\pm$4E-3\\
Wine 			&0.64$\pm$0.01&\bf\color{teal}0.63$\pm$0.01&\bf\color{orange}0.62$\pm$0.01&\bf\color{teal}0.63$\pm$0.04&-0.97$\pm$0.01&\bf\color{teal}-0.95$\pm$0.01&\bf\color{orange}-0.93$\pm$0.06&-0.96$\pm$0.06\\
Yacht 			&1.02$\pm$0.05&\bf\color{orange}0.71$\pm$0.05&1.11$\pm$0.38&\bf\color{teal}1.02$\pm$0.42&-1.63$\pm$0.02&\bf\color{orange}-1.30$\pm$0.02&-1.55$\pm$0.12&\bf\color{teal}{-1.49$\pm$0.45}\\
\bottomrule
\end{tabular}}\label{TAB:Benchmark_reg}
\end{table*}

The results reported in Table \ref{TAB:Benchmark_reg} indicate that TAGI matches the performance of existing methods in terms of root mean square error (RMSE) and log-likelihood (LL).  Even though VMG displays the best predictive performance, its computational time is two orders of magnitude greater than TAGI, PBP and MC-Dropout \citep{sun2017learning}. The timing details reported in Appendix \ref{Appendix:C} show that the current TAGI's implementation has a computational time that is more than four times faster than PBP and MC-Dropout \citep{gal2016dropout}. Moreover, because TAGI had to be implemented from scratch, it is not yet fully optimized for computational efficiency. The same is true for the optimization of hyper-parameters such as $\sigma_{V}$, which is assumed to be constant over the covariate domain, and which is currently treated as an hyperparameter to be estimated separately from the analytical inference for weights and biases. The fact that TAGI is currently limited to the case of homoscedastic variance reduces its performance in comparison with the other methods presented, and also inevitably lead to poorly calibrated predictive uncertainties. Note that as it was the case for the other methods reported in Table \ref{TAB:Benchmark_reg}, the number of epochs employed was not optimized.

\subsection{Application on MNIST}\label{S:MNIST}
We apply TAGI to the MNIST classification problem \citep{lecun1998gradient} consisting of $\mathtt{D}=70\,000$ $(28\times28)$ greyscale images for $\mathtt{K}=10$ classes ($60\,000$ training and $10\,000$ test). Here, we compare the performance of two AG-FNN configurations, each having $\mathtt{L}=2$ hidden layers with a number of hidden units equal to $\mathtt{A}\in\{100, 800\}$. Each AG-FNN has the same structure for the input ($\mathtt{X}=784$ nodes) and the output layer ($\mathtt{Y}=11$ nodes). The ReLU activation function is used for the two hidden layers. For each digit, the vector of covariates $\mathbf{x}_{i}\in(0,1)^{784}$ is assumed to be deterministic so that $\bm{\mu}_{\bm{X}_{i}}=\mathbf{x}_{i}$ and $\mathbf{\Sigma}_{\bm{X}_{i}}=\mathbf{0}$. The prior covariance for bias is initialized to $\Sigma_{\bm{B}}^{\texttt{0}}=0.01\cdot\mathbf{I}$, and by using the Xaviers's initialization approach \citep{glorot2010understanding} for weights $\Sigma_{\bm{W}}^{\texttt{0}}$. The prior mean vector is randomly sampled from $\bm{\mu}_{\bm{\theta}}^{\texttt{0}}\sim\mathcal{N}(\mathbf{0},\Sigma_{\bm{\theta}}^{0})$.
The hyper-parameter associated with the output layer is set to $\alpha=1/3$. The posterior mean vector as well as the main diagonal of the posterior covariance are learnt using two setups: (1) a single observation per batch, that is, $\mathtt{B}=1$, and (2) $\mathtt{B}=10$ observations per batch. Each network is evaluated for $\sigma_{V} = \{0.1, 0.2, 0.3, 0.4\}$ and the optimal value for $\sigma_{V}$ is selected using a randomly selected validation set corresponding to 5\% of the training set. The optimal number of epochs $\mathtt{E}$ is identified using an early-stop procedure evaluated on the validation set.

Table \ref{T:TE} presents the average test error evaluated on $10\,000$ images for the different AG-FNN configurations for $\mathtt{E}=1$ epoch, and for the optimal number of epochs found using early stopping. In order to factor in the effect of random weight initialization, the results reported are the average and standard deviations from five runs.\begin{table}[h!]\small 
\caption{\small MNIST test-set average classification error [\%] for the first and last epoch. $\mathtt{A}$: number of hidden units on each layer; $\mathtt{B}$: number of observations per batch; $\mathtt{E}$: number of epochs; $e$: optimal number of epoch found using early-stop. The results reported are the average of five runs along with $\pm$ one standard deviation.}
\centering\scalebox{1}{
\begin{tabular}{ccccc|cccc}    	\addlinespace
    \toprule
    	\multicolumn{1}{l}{}&\multicolumn{4}{c}{$\mathtt{B}=1$} &\multicolumn{4}{c}{$\mathtt{B}=10$} \\[0pt] 
	\cmidrule{2-9}
$\mathtt{A}$&$\mathtt{E}=1$& $\mathtt{E}=e$ &$e$&$\sigma_V$&$\mathtt{E}=1$& $\mathtt{E}=e$ &$e$&$\sigma_V$\\[0pt]   
\cmidrule{1-9}
\multirow{1}{*}{$100$}&$3.39\!\pm\!0.17$&$2.29\!\pm\!0.06$&$21\!\pm\!12$&$0.2$&$3.56\!\pm\!0.2$&$2.45\!\pm\!0.20$&$20\!\pm\!12$&$0.4$\\[2pt]
$800$&$3.15\!\pm\!0.2$ &$1.54\!\pm\!0.07$&$14\!\pm\!9$&$0.1$&$3.10\!\pm\!0.2$&$1.53\!\pm\!0.07$&$15\!\pm\!6$&$0.4$\\[2pt]
\bottomrule
\end{tabular}}
\label{T:TE}
\end{table} The performance achieved with respect to the average classification errors matches the reported state-of-the-art accuracy of approximately 1.6\% for FNNs having a same architecture with 2 layers and 800 hidden units and trained using gradient backpropagation \citep{simard2003best,pmlr-v28-wan13}. Some Bayesian methods for neural networks have shown to be able to outperform their deterministic counterparts. For example,  \cite{blundell2015weight} showed that they can reach an error rate of 1.34\% using Bayes by Backprop with scale mixture, while only reaching 1.99\% when using Gaussians. Similarly \cite{louizos2016structured} showed that VMG could reach a performance of 1.15\% with even smaller networks. The results in Table \ref{T:TE} indicate that TAGI's classification accuracy is not significantly affected by the usage of batch sizes greater than one. Nevertheless, we noticed though our experiments that using large batch sizes makes the learning phase sensitive to the network initialization as well as the observation noise parameter $\sigma_{V}$.

\section{Conclusion} \label{S:Conclusion}
The tractable approximate Gaussian inference method proposed in this paper allows for the analytical inference of the posterior mean vector and diagonal covariance matrix for the parameters of Bayesian neural networks. The applications on the regression and classification datasets validate that the approach matches the performance of existing methods with respect to computational efficiency and accuracy. TAGI's performance and its linear complexity with respect to the number of parameters makes it a viable alternative to gradient backpropagation.  By allowing the  treatment of uncertainty and by being inherently suited for online inference, we foresee that the approach will enable transformative developments in supervised, unsupervised, and reinforcement learning. 

\bigskip

\section*{Acknowledgements}
The second author was financially supported by research grants from Hydro-Quebec/IREQ, and the Natural Sciences and Engineering Research Council of Canada (NSERC). The third author was supported by a research grant from the Institute for Data Valorization (IVADO).
\bibliography{Goulet_reference_librairy}

\begin{thebibliography}{37}
\providecommand{\natexlab}[1]{#1}
\providecommand{\url}[1]{\texttt{#1}}
\expandafter\ifx\csname urlstyle\endcsname\relax
  \providecommand{\doi}[1]{doi: #1}\else
  \providecommand{\doi}{doi: \begingroup \urlstyle{rm}\Url}\fi

\bibitem[Arasaratnam and Haykin(2008)]{arasaratnam2008nonlinear}
I.~Arasaratnam and S.~Haykin.
\newblock Nonlinear {B}ayesian filters for training recurrent neural networks.
\newblock In \emph{Mexican International Conference on Artificial
  Intelligence}, pages 12--33. Springer, 2008.

\bibitem[Barber and Bishop(1998)]{barber1998ensemble}
D.~Barber and C.~M. Bishop.
\newblock Ensemble learning in {B}ayesian neural networks.
\newblock \emph{NATO ASI Series F Computer and Systems Sciences}, 168:\penalty0
  215--238, 1998.

\bibitem[Blundell et~al.(2015)Blundell, Cornebise, Kavukcuoglu, and
  Wierstra]{blundell2015weight}
C.~Blundell, J.~Cornebise, K.~Kavukcuoglu, and D.~Wierstra.
\newblock Weight uncertainty in neural network.
\newblock In \emph{{International Conference on Machine Learning}}, pages
  1613--1622, 2015.

\bibitem[Cobb et~al.(2019)Cobb, Baydin, Markham, and
  Roberts]{cobb2019introducing}
A.D. Cobb, At{\i}l{\i}m~G. Baydin, A.~Markham, and S.J. Roberts.
\newblock Introducing an explicit symplectic integration scheme for
  {R}iemannian manifold {H}amiltonian {M}onte {C}arlo.
\newblock \emph{arXiv preprint arXiv:1910.06243}, 2019.

\bibitem[Efron(2012)]{efron2012large}
B.~Efron.
\newblock \emph{Large-scale inference: empirical Bayes methods for estimation,
  testing, and prediction}, volume~1.
\newblock Cambridge University Press, 2012.

\bibitem[Farquhar and Gal(2019)]{farquhar2019unifying}
S.~Farquhar and Y.~Gal.
\newblock A unifying {B}ayesian view of continual learning.
\newblock \emph{arXiv preprint arXiv:1902.06494}, 2019.

\bibitem[Farquhar et~al.(2020)Farquhar, Smith, and Gal]{NEURIPS2020_2dfe1946}
S.~Farquhar, L.~Smith, and Y.~Gal.
\newblock Liberty or depth: Deep bayesian neural nets do not need complex
  weight posterior approximations.
\newblock In H.~Larochelle, M.~Ranzato, R.~Hadsell, M.~F. Balcan, and H.~Lin,
  editors, \emph{Advances in Neural Information Processing Systems}, volume~33,
  pages 4346--4357, 2020.

\bibitem[Gal and Ghahramani(2016)]{gal2016dropout}
Y.~Gal and Z.~Ghahramani.
\newblock Dropout as a {B}ayesian approximation: Representing model uncertainty
  in deep learning.
\newblock In \emph{{International Conference on Machine Learning}}, pages
  1050--1059, 2016.

\bibitem[Gelman et~al.(2014)Gelman, Carlin, Stern, and
  Rubin]{gelman2014bayesian}
A.~Gelman, J.~B. Carlin, H.~S. Stern, and D.~B. Rubin.
\newblock \emph{Bayesian data analysis}.
\newblock CRC Press, 3rd edition, 2014.

\bibitem[Ghahramani(2015)]{ghahramani2015probabilistic}
Z.~Ghahramani.
\newblock Probabilistic machine learning and artificial intelligence.
\newblock \emph{Nature}, 521\penalty0 (7553):\penalty0 452, 2015.

\bibitem[Glorot and Bengio(2010)]{glorot2010understanding}
X.~Glorot and Y.~Bengio.
\newblock Understanding the difficulty of training deep feedforward neural
  networks.
\newblock In \emph{Proceedings of the thirteenth {International Conference on
  Artificial Intelligence and Statistics}}, pages 249--256, 2010.

\bibitem[Goodfellow et~al.(2016)Goodfellow, Bengio, and
  Courville]{goodfellow2016deep}
I.~Goodfellow, Y.~Bengio, and A.~Courville.
\newblock \emph{Deep learning}.
\newblock MIT Press, 2016.

\bibitem[Haykin(2004)]{haykin2004kalman}
S.~Haykin.
\newblock \emph{Kalman filtering and neural networks}, volume~47.
\newblock John Wiley \& Sons, 2004.

\bibitem[Hern{\'a}ndez-Lobato and Adams(2015)]{hernandez2015probabilistic}
J.~M. Hern{\'a}ndez-Lobato and R.~Adams.
\newblock Probabilistic backpropagation for scalable learning of {B}ayesian
  neural networks.
\newblock In \emph{{International Conference on Machine Learning}}, pages
  1861--1869, 2015.

\bibitem[Hinton and Van~Camp(1993)]{hinton1993keeping}
G.~E. Hinton and D.~Van~Camp.
\newblock Keeping the neural networks simple by minimizing the description
  length of the weights.
\newblock In \emph{Proceedings of the sixth annual conference on Computational
  learning theory}, pages 5--13. ACM, 1993.

\bibitem[Jin et~al.(2018)Jin, Keutzer, and Levine]{pmlr-v80-jin18c}
P.~Jin, K.~Keutzer, and S.~Levine.
\newblock Regret minimization for partially observable deep reinforcement
  learning.
\newblock In Jennifer Dy and Andreas Krause, editors, \emph{Proceedings of the
  35th International Conference on Machine Learning}, volume~80 of
  \emph{Proceedings of Machine Learning Research}, pages 2342--2351, 10--15 Jul
  2018.

\bibitem[Kendall and Gal(2017)]{kendall2017uncertainties}
A.~Kendall and Y.~Gal.
\newblock What uncertainties do we need in {B}ayesian deep learning for
  computer vision?
\newblock In \emph{Advances in neural information processing systems}, pages
  5574--5584, 2017.

\bibitem[Kingma et~al.(2015)Kingma, Salimans, and Welling]{NIPS2015_bc731692}
D.~P. Kingma, T.~Salimans, and M.~Welling.
\newblock Variational dropout and the local reparameterization trick.
\newblock In C.~Cortes, N.~Lawrence, D.~Lee, M.~Sugiyama, and R.~Garnett,
  editors, \emph{Advances in Neural Information Processing Systems}, volume~28,
  2015.

\bibitem[LeCun et~al.(1998)LeCun, Bottou, Bengio, and
  Haffner]{lecun1998gradient}
Y.~LeCun, L.~Bottou, Y.~Bengio, and P.~Haffner.
\newblock Gradient-based learning applied to document recognition.
\newblock \emph{Proceedings of the IEEE}, 86\penalty0 (11):\penalty0
  2278--2324, 1998.

\bibitem[Louizos and Welling(2016)]{louizos2016structured}
C.~Louizos and M.~Welling.
\newblock Structured and efficient variational deep learning with matrix
  {Gaussian} posteriors.
\newblock In \emph{{International Conference on Machine Learning}}, pages
  1708--1716, 2016.

\bibitem[MacKay(1992)]{mackay1992practical}
D.J.C. MacKay.
\newblock A practical {B}ayesian framework for backpropagation networks.
\newblock \emph{Neural computation}, 4\penalty0 (3):\penalty0 448--472, 1992.

\bibitem[Morin and Bengio(2005)]{morin2005hierarchical}
F.~Morin and Y.~Bengio.
\newblock Hierarchical probabilistic neural network language model.
\newblock In \emph{Aistats}, volume~5, pages 246--252. Citeseer, 2005.

\bibitem[Murphy(2012)]{murphy2012machine}
K.~P. Murphy.
\newblock \emph{Machine learning: a probabilistic perspective}.
\newblock The MIT Press, 2012.

\bibitem[Neal(1995)]{neal1995Bayesian}
R.~M. Neal.
\newblock \emph{{B}ayesian learning for neural networks}.
\newblock PhD thesis, University of Toronto, 1995.

\bibitem[Osawa et~al.(2019)Osawa, Swaroop, Jain, Eschenhagen, Turner, Yokota,
  and Khan]{Osawa2019PracticalDL}
K.~Osawa, S.~Swaroop, A.~Jain, R.~Eschenhagen, R.~E. Turner, R.~Yokota, and
  M.~E. Khan.
\newblock Practical deep learning with {B}ayesian principles.
\newblock In \emph{Advances in Neural Information Processing Systems}, 2019.

\bibitem[Plumer(1995)]{plumer1995training}
E.~S. Plumer.
\newblock Training neural networks using sequential extended {Kalman}
  filtering.
\newblock In \emph{1995 world congress on neural networks}. Los Alamos National
  Lab., NM (United States), 1995.

\bibitem[Puskorius and Feldkamp(1991)]{puskorius1991decoupled}
G.~V. Puskorius and L.~A. Feldkamp.
\newblock Decoupled extended {K}alman filter training of feedforward layered
  networks.
\newblock In \emph{{IJCNN-91-Seattle International Joint Conference on Neural
  Networks}}, volume~1, pages 771--777. IEEE, 1991.

\bibitem[Rasmussen and Williams(2006)]{williams2006gaussian}
C.~E. Rasmussen and C.~K. Williams.
\newblock \emph{Gaussian processes for machine learning}.
\newblock the MIT Press, 2006.

\bibitem[Rauch et~al.(1965)Rauch, Striebel, and Tung]{rauch1965maximum}
H.~E. Rauch, C.~T. Striebel, and F.~Tung.
\newblock Maximum likelihood estimates of linear dynamic systems.
\newblock \emph{AIAA journal}, 3\penalty0 (8):\penalty0 1445--1450, 1965.

\bibitem[Rumelhart et~al.(1986)Rumelhart, Hinton, and
  Williams]{rumelhart1986learning}
D.~E. Rumelhart, G.~E. Hinton, and R.~J. Williams.
\newblock Learning representations by back-propagating errors.
\newblock \emph{nature}, 323\penalty0 (6088):\penalty0 533, 1986.

\bibitem[Simard et~al.(2003)Simard, Steinkraus, and Platt]{simard2003best}
P.~Y Simard, D.~Steinkraus, and J.~C. Platt.
\newblock Best practices for convolutional neural networks applied to visual
  document analysis.
\newblock In \emph{{Proceedings of Seventh International Conference on Document
  Analysis and Recognition}}, 2003.

\bibitem[Singhal and Wu(1989)]{NIPS1988_101}
S.~Singhal and L.~Wu.
\newblock Training multilayer perceptrons with the extended {K}alman algorithm.
\newblock In D.~S. Touretzky, editor, \emph{Advances in Neural Information
  Processing Systems 1}, pages 133--140. Morgan-Kaufmann, 1989.

\bibitem[Sun et~al.(2017)Sun, Chen, and Carin]{sun2017learning}
S.~Sun, C.~Chen, and L.~Carin.
\newblock Learning structured weight uncertainty in {B}ayesian neural networks.
\newblock In \emph{Artificial Intelligence and Statistics}, pages 1283--1292,
  2017.

\bibitem[Sutton and Barto(2018)]{sutton2011reinforcement}
R.~S. Sutton and A.~G. Barto.
\newblock \emph{Reinforcement learning: An introduction}.
\newblock MIT Press, 2nd edition, 2018.

\bibitem[Wan and Merwe(2000)]{wan2000unscented}
E.~A. Wan and R.~v.~d. Merwe.
\newblock The unscented {Kalman} filter for nonlinear estimation.
\newblock In \emph{Adaptive Systems for Signal Processing, Communications, and
  Control Symposium}, pages 153--158. IEEE, 2000.

\bibitem[Wan et~al.(2013)Wan, Zeiler, Zhang, Cun, and Fergus]{pmlr-v28-wan13}
L.~Wan, M.~Zeiler, S.~Zhang, Y.~Le Cun, and R.~Fergus.
\newblock Regularization of neural networks using {DropConnect}.
\newblock In \emph{Proceedings of the 30th International Conference on Machine
  Learning}, volume~28 of \emph{Proceedings of Machine Learning Research},
  pages 1058--1066, 2013.

\bibitem[Wu et~al.(2019)Wu, Nowozin, Meeds, Turner, Hern{\'{a}}ndez{-}Lobato,
  and Gaunt]{DBLP:confWuNMTHG19}
A.~Wu, S.~Nowozin, E.~Meeds, R.~E. Turner, J.~M. Hern{\'{a}}ndez{-}Lobato, and
  A.~L. Gaunt.
\newblock Deterministic variational inference for robust {B}ayesian neural
  networks.
\newblock In \emph{7th International Conference on Learning Representations,
  {ICLR} 2019, New Orleans, LA, USA, May 6-9, 2019}, 2019.

\end{thebibliography}
%
%
%
\newpage
\appendix
%
%
\section{Feedforward neural network nomenclature}\label{A:FNN}
\begin{figure}[htbp]\centering
\subfloat[][Expanded representation]{\tikzset{dist1/.style={path picture= {
    \begin{scope}[x=1pt,y=10pt]
      \draw plot[domain=-6:6] (\x,{1/(1 + exp(-\x))-0.5});
    \end{scope}
    }
  }
}
\tikzset{dist2/.style={path picture= {
    \begin{scope}[x=1pt,y=10pt]
      \draw plot[domain=-6:6] (\x,{\x/10});
    \end{scope}
    }
  }
}
\tikzstyle{input}=[draw,fill=red!10,circle,minimum size=15pt,inner sep=0pt]
\tikzstyle{hidden}=[draw,fill=green!20,circle,minimum size=15pt,inner sep=0pt]
\tikzstyle{output}=[draw,fill=blue!20,circle,minimum size=15pt,inner sep=0pt]
\tikzstyle{bias}=[draw=none,circle,minimum size=15pt,inner sep=0pt]

\tikzstyle{stateTransition}=[->,line width=0.25pt,draw=black!50]

\begin{tikzpicture}[scale=1.36]\scriptsize
    \node (x1)[input]   at (0, 1.25) {$x_{1}$};
    \node (x2)[input] at (0, 0.25) {$x_{2}$};
    \node (x3)  at (0,-0.5) {\raisebox{6pt}{\vdots}};
    \node (xx)[input]  at (0,-1.25) {$x_{\mathtt{X}}$};

    \node (b0)[bias]  at (0.35,1.9) {$\bm{b}^{(\texttt{0})}$};

    \node (h11)[hidden] at (1, 1.5) {$z^{(1)}_{1}$};
    \node (h12)[hidden] at (1, 0.5) {$z^{(1)}_{2}$};
    \node (h13) at (1,-0.5) {\raisebox{5pt}{$\vdots$}};
    \node (h14)[hidden] at (1,-1.5) {$z^{(1)}_{\mathtt{A}}$};
    \node (b1)[bias]  at (1.35,1.9) {$\bm{b}^{(1)}$};

    \node (h21)[hidden] at (2, 1.5) {$z^{(2)}_{1}$};
    \node (h22)[hidden] at (2, 0.5) {$z^{(2)}_{2}$};
    \node (h23) at (2,-0.5) {\raisebox{6pt}{\vdots}};
    \node (h24)[hidden] at (2,-1.5) {$z^{(2)}_{\mathtt{A}}$};
    \node (b2)[bias]  at (2.35,1.9) {$\bm{b}^{(\cdots)}$};
    
    \node (h31) at (3, 1.5) {$\cdots$};
    \node (h32) at (3, 0.5) {$\cdots$};
    \node (h33) at (3,-0.5) {\raisebox{6pt}{\vdots}};
    \node (h34) at (3,-1.5) {$\cdots$};
    \node (b3)[bias]  at (3.35,1.9) {$\bm{b}^{(\mathtt{L}-1)}$};
    
    \node (h41)[hidden] at (4, 1.5) {$z^{(\mathtt{L})}_{1}$};
    \node (h42)[hidden] at (4, 0.5) {$z^{(\mathtt{L})}_{2}$};
    \node (h43) at (4,-0.5) {\raisebox{6pt}{\vdots}};
    \node (h44)[hidden] at (4,-1.5) {$z^{(\mathtt{L})}_{\mathtt{A}}$};
    \node (b4)[bias]  at (4.35,1.9) {$\bm{b}^{(\mathtt{L})}$};

    \node (h1)[hidden] at (5.1,1.25) {$z_{1}^{(\mathtt{O})}$};
    \node (h2)[hidden] at (5.1,0.25) {$z_{2}^{(\mathtt{O})}$};
    \node (hn) at (5.1,-0.5) {\raisebox{6pt}{\vdots}};
    \node (h3)[hidden] at (5.1,-1.25) {$z_{\mathtt{Y}}^{(\mathtt{O})}$};
    \node (o1)[output] at (5.7,1.25) {$y_{1}$};
    \node (o2)[output] at (5.7,0.25) {$y_{2}$};
    \node (on) at (5.7,-0.5) {\raisebox{6pt}{\vdots}};
    \node (o3)[output] at (5.7,-1.25) {$y_{\mathtt{Y}}$};

    \draw[stateTransition] (x1) -- (h11) node  {};
    \draw[stateTransition] (x1) -- (h12) node  {};
    \draw[stateTransition] (x1) -- (h13) node  {};
    \draw[stateTransition] (x1) -- (h14) node  {};
    \draw[stateTransition] (x2) -- (h11) node  {};
    \draw[stateTransition] (x2) -- (h12) node  {};
    \draw[stateTransition] (x2) -- (h13) node  {};
    \draw[stateTransition] (x2) -- (h14) node  {};
    \draw[stateTransition] (x3) -- (h11) node  {};
    \draw[stateTransition] (x3) -- (h12) node  {};
    \draw[stateTransition] (x3) -- (h13) node  {};
    \draw[stateTransition] (x3) -- (h11) node  {};
    \draw[stateTransition] (xx) -- (h11) node  {};
    \draw[stateTransition] (xx) -- (h12) node  {};
    \draw[stateTransition] (xx) -- (h13) node  {};
    \draw[stateTransition] (xx) -- (h11) node  {};
     \draw[stateTransition] (xx) -- (h14) node  {};
    
    \draw[stateTransition] (b0) -- (h12) node {};
    \draw[stateTransition] (b0) -- (h11) node {};
    \draw[stateTransition] (b0) -- (h13) node {};
    \draw[stateTransition] (b0) -- (h14) node {};

    \draw[stateTransition,opacity=1] (h11) -- (h21) node [midway, rotate=0,fill=white,opacity=0.8] {\scriptsize$w^{(1)}_{1,1}$};
    \draw[stateTransition,opacity=0] (h11) -- (h21) node [midway, rotate=0,fill=white] {\scriptsize$w^{(1)}_{1,1}$};
    \draw[stateTransition,opacity=0.2] (h11) -- (h22) node  {};
    \draw[stateTransition,opacity=0.2] (h11) -- (h23) node  {};
    \draw[stateTransition,opacity=0.2] (h11) -- (h24) node  {};
    \draw[stateTransition,opacity=1] (h12) -- (h21) node [midway, rotate=45,fill=white,opacity=0.8] {\scriptsize$w^{(1)}_{1,2}$};
        \draw[stateTransition,opacity=0] (h12) -- (h21) node [midway, rotate=45,fill=white!90] {\scriptsize$w^{(1)}_{2,1}$};
    \draw[stateTransition,opacity=0.2] (h12) -- (h22) node  {};
    \draw[stateTransition,opacity=0.2] (h12) -- (h23) node  {};
    \draw[stateTransition,opacity=0.2] (h12) -- (h24) node  {};
        \draw[stateTransition,opacity=1] (h13) -- (h21) node [midway, rotate=60,fill=white,opacity=0.8] {\scriptsize$w^{(1)}_{1,i}$};
     \draw[stateTransition,opacity=0] (h13) -- (h21) node [midway, rotate=60] {\scriptsize$w^{(1)}_{i,1}$};
    \draw[stateTransition,opacity=0.2] (h13) -- (h22) node  {};
    \draw[stateTransition,opacity=0.2] (h13) -- (h23) node  {};
    \draw[stateTransition,opacity=0.2] (h13) -- (h24) node  {};
    \draw[stateTransition,opacity=1] (h14) -- (h21) node [midway, rotate=70,fill=white,opacity=0.8] {\scriptsize$w^{(1)}_{1,\mathtt{A}}$};
     \draw[stateTransition,opacity=0] (h14) -- (h21) node [midway, rotate=70] {\scriptsize$w^{(1)}_{\mathtt{A},1}$};
    \draw[stateTransition,opacity=0.2] (h14) -- (h22) node  {};
    \draw[stateTransition,opacity=0.2] (h14) -- (h23) node  {};
    \draw[stateTransition,opacity=0.2] (h14) -- (h24) node  {};

    \draw[stateTransition,opacity=1] (b1) -- (h21) node  [midway,above=-0.05cm, rotate=-25] {\scriptsize$b^{(1)}_{1}$};
    \draw[stateTransition,opacity=0.2] (b1) -- (h22) node {};
    \draw[stateTransition,opacity=0.2] (b1) -- (h23) node {};
    \draw[stateTransition,opacity=0.2] (b1) -- (h24) node {};
    
     \draw[stateTransition] (h21) -- (h31) node  {};
    \draw[stateTransition] (h21) -- (h32) node  {};
    \draw[stateTransition] (h21) -- (h33) node  {};
    \draw[stateTransition] (h21) -- (h34) node  {};
    \draw[stateTransition] (h22) -- (h31) node  {};
    \draw[stateTransition] (h22) -- (h32) node  {};
    \draw[stateTransition] (h22) -- (h33) node  {};
    \draw[stateTransition] (h22) -- (h34) node  {};
    \draw[stateTransition] (h23) -- (h31) node  {};
    \draw[stateTransition] (h23) -- (h32) node  {};
    \draw[stateTransition] (h23) -- (h33) node  {};
    \draw[stateTransition] (h23) -- (h34) node  {};
    \draw[stateTransition] (h24) -- (h31) node  {};
    \draw[stateTransition] (h24) -- (h32) node  {};
    \draw[stateTransition] (h24) -- (h33) node  {};
    \draw[stateTransition] (h24) -- (h31) node  {};
    \draw[stateTransition] (h24) -- (h34) node  {};

    \draw[stateTransition] (b2) -- (h32) node {};
    \draw[stateTransition] (b2) -- (h31) node {};
    \draw[stateTransition] (b2) -- (h33) node {};
    \draw[stateTransition] (b2) -- (h34) node {};
    
    \draw[stateTransition] (h31) -- (h41) node  {};
    \draw[stateTransition] (h31) -- (h42) node  {};
    \draw[stateTransition] (h31) -- (h43) node  {};
    \draw[stateTransition] (h31) -- (h44) node  {};
    \draw[stateTransition] (h32) -- (h41) node  {};
    \draw[stateTransition] (h32) -- (h42) node  {};
    \draw[stateTransition] (h32) -- (h43) node  {};
    \draw[stateTransition] (h32) -- (h44) node  {};
    \draw[stateTransition] (h33) -- (h41) node  {};
    \draw[stateTransition] (h33) -- (h42) node  {};
    \draw[stateTransition] (h33) -- (h43) node  {};
    \draw[stateTransition] (h33) -- (h44) node  {};
    \draw[stateTransition] (h34) -- (h41) node  {};
    \draw[stateTransition] (h34) -- (h42) node  {};
    \draw[stateTransition] (h34) -- (h43) node  {};
    \draw[stateTransition] (h34) -- (h41) node  {};
     \draw[stateTransition] (h34) -- (h44) node  {};
    
    \draw[stateTransition] (b3) -- (h42) node {};
    \draw[stateTransition] (b3) -- (h41) node {};
    \draw[stateTransition] (b3) -- (h43) node {};
    \draw[stateTransition] (b3) -- (h44) node {};

    \draw[stateTransition] (h41) -- (h1) node  {};
    \draw[stateTransition] (h42) -- (h1) node  {};
    \draw[stateTransition] (h43) -- (h1) node  {};
    \draw[stateTransition] (h44) -- (h1) node  {};
    \draw[stateTransition] (b4) -- (h1) node {};
    
    \draw[stateTransition] (h41) -- (h2) node  {};
    \draw[stateTransition] (h42) -- (h2) node  {};
    \draw[stateTransition] (h43) -- (h2) node  {};
    \draw[stateTransition] (h44) -- (h2) node  {};
    \draw[stateTransition] (b4) -- (h2) node {};
    
    \draw[stateTransition] (h41) -- (hn) node  {};
    \draw[stateTransition] (h42) -- (hn) node  {};
    \draw[stateTransition] (h43) -- (hn) node  {};
    \draw[stateTransition] (h44) -- (hn) node  {};
    \draw[stateTransition] (b4) -- (hn) node {};
    
    \draw[stateTransition] (h41) -- (h3) node  {};
    \draw[stateTransition] (h42) -- (h3) node  {};
    \draw[stateTransition] (h43) -- (h3) node  {};
    \draw[stateTransition] (h44) -- (h3) node  {};
    \draw[stateTransition] (b4) -- (h3) node {};
    
     \draw[stateTransition] (h1) -- (o1) node  {};
     \draw[stateTransition] (h2) -- (o2) node  {};
     \draw[stateTransition] (hn) -- (on) node  {};
     \draw[stateTransition] (h3) -- (o3) node  {};
\end{tikzpicture}}\\\subfloat[][Compact representation where $\bm{\theta}=\{\bm{w},\bm{b}\}$]{{\tikzset{dist1/.style={path picture= {
    \begin{scope}[x=1pt,y=10pt]
      \draw plot[domain=-6:6] (\x,{1/(1 + exp(-\x))-0.5});
    \end{scope}
    }
  }
}
\tikzset{dist2/.style={path picture= {
    \begin{scope}[x=1pt,y=10pt]
      \draw plot[domain=-6:6] (\x,{\x/10});
    \end{scope}
    }
  }
}
\tikzstyle{input}=[draw,fill=red!10,circle,minimum size=15pt,inner sep=0pt]
\tikzstyle{hidden}=[draw,fill=green!20,circle,minimum size=15pt,inner sep=0pt]
\tikzstyle{output}=[draw,fill=blue!20,circle,minimum size=15pt,inner sep=0pt]
\tikzstyle{bias}=[draw=none,circle,minimum size=15pt,inner sep=0pt]

\tikzstyle{stateTransition}=[->,line width=0.25pt,draw=black!50]

\begin{tikzpicture}[scale=1.36]\scriptsize
    \node (x)[input]   at (0, 1.5) {$\bm{x}$};
    \node (z1)[hidden] at (1, 1.5) {$\,\bm{z}^{\!(1)}$};
    \node (z2)[hidden] at (2, 1.5) {$\,\bm{z}^{\!(2)}$};
    \node (zi) at (3, 1.5) {$\cdots$};
    \node (zL)[hidden] at (4, 1.5) {$\,\bm{z}^{\!(\mathtt{L})}$};
    \node (zO)[hidden] at (5.1,1.5) {$\,\bm{z}^{\!(\mathtt{O})}$};

    \node (y)[output] at (5.7,1.5) {$\bm{y}$};

    \draw[stateTransition,opacity=1] (x) -- (z1) node [midway, rotate=0,fill=white,opacity=0.8] {\scriptsize$\!\bm{\theta}^{(\texttt{0})}\!$};
    \draw[stateTransition,opacity=1] (z1) -- (z2) node [midway, rotate=0,fill=white,opacity=0.8] {\scriptsize$\!\bm{\theta}^{(1)}\!$};
     \draw[stateTransition,opacity=1] (z2) -- (zi) node [midway, rotate=0,fill=white,opacity=0.8] {\scriptsize$\!\bm{\theta}^{(2)}\!$};
     \draw[stateTransition,opacity=1] (zi) -- (zL) node [midway, rotate=0,fill=white,opacity=0.8] {\scriptsize$\!\!\bm{\theta}^{(\mathtt{L}\text{-}1)}\!\!$};
    \draw[stateTransition,opacity=1] (zL) -- (zO) node [midway, rotate=0,fill=white,opacity=0.8] {\scriptsize$\!\bm{\theta}^{(\mathtt{L})}\!$};
     \draw[stateTransition,opacity=1] (zO) -- (y);
\end{tikzpicture}}}
\caption{Expanded and compact representations of the variable nomenclature associated with feedforward neural networks.}
\label{FIG:FNN}
\end{figure}
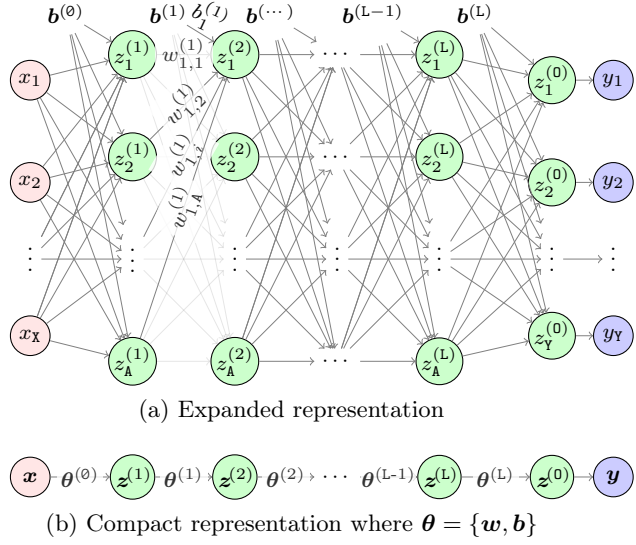
\section{Derivation for the Gaussian multiplicative approximation }\label{Appendix:A}
The proof of statement \eqref{eq1} can be obtained directly from the definition of covariance.  To prove  the statement  \eqref{eq2}, one needs the moments of variables and their products. Because the underlying distribution is Gaussian and its moment generating function are known, $M_{\bm{X}}(\mathbf{t}^{\intercal})=\mathbb{E}[e^{\mathbf{t}^{\intercal}\bm{X}}]=e^{\mathbf{t}^{\intercal}\bm{\mu}+\frac{1}{2}\mathbf{t}^{\intercal}\bm{\Sigma} \mathbf{t}}, \mathbf{t}=[t_1~\ldots~t_4]^{\intercal}$, all moments can be extracted using the moment-generating function. 
Using the derivatives of the moment-generating function, 
 \begin{eqnarray}
\mathbb{E}[X_1X_2X_3]&=&\frac{\partial^3}{\partial t_1 \partial t_2 \partial t_3} \mathbb{E}[e^{\mathbf{t}^{\intercal}\bm{x}}] |_{t_1 =t_2=t_3=t_4=0 }\nonumber\\
&=&
  \frac{\partial^3}{\partial t_1 \partial t_2 \partial t_3}  e^{\mathbf{t}^{\intercal}\bm{\mu}+\frac{1}{2}\mathbf{t}^{\intercal}\mathbf{\Sigma} \mathbf{t}}|_{t_1 =t_2=t_3=t_4=0} \nonumber\\
&&\!\!\!\!\!\!\!\!\!\!\!\!\!\!\!\!\!\!\!\!\!\!\!\!\!\!\!\!\!\!\!\!\!\!\!\!\!\!\!\!\!\!\!\!\!\!= \frac{\partial^3}{\partial t_1 \partial t_2 \partial t_3}  e^{\sum_{i=1}^{4}\!t_i \mu_i+\frac{1}{2} \sum_{i,j=1}^{4}\!t_it_j \text{cov}(X_i,X_j) }|_{t_{1,2,3,4}=0} \nonumber\\
&&\!\!\!\!\!\!\!\!\!\!\!\!\!\!\!\!\!\!\!\!\!\!\!\!\!\!\!\!\!\!\!\!\!\!\!\!\!\!\!\!\!\!\!\!\!\!= \text{cov}(X_1,X_2)\mu_3+\text{cov}(X_1,X_3)\mu_2\nonumber\\&&\!\!\!\!\!\!\!\!\!\!\!\!\!\!\!\!\!\!\!\!\!\!\!\!+\text{cov}(X_2,X_3)\mu_1+\mu_1\mu_2\mu_3\label{E123}.
 \end{eqnarray}
 Using the definition of covariance and substituting \eqref{E123}, the statement  \eqref{eq2} can be established: 
  \begin{eqnarray}
 \text{cov}(X_3,X_1X_2)&=&\mathbb{E}[X_1X_2X_3]-\mathbb{E}(X_1X_3)\mathbb{E}(X_2)\nonumber\\&=& \text{cov}(X_1,X_2)\mu_3+\text{cov}(X_1,X_3)\mu_2\nonumber \\ &&
 +\text{cov}(X_2,X_3)\mu_1+\mu_1\mu_2\mu_3 \nonumber\\ && - \mu_3 \big(\mu_1\mu_2+\text{cov}(X_1,X_2)\big).
\end{eqnarray}
The expansion of the right side for the last statement leads to Equation \eqref{eq3} so that 
$$
\text{cov}(X_1X_2,X_3X_4)=\mathbb{E}[X_1X_2X_3X_4]-\mathbb{E}[X_1X_2]\mathbb{E}[X_3X_4],
$$
where the expected value of  the product two random variables is given in \eqref{eq1} and the expectation for the product of four random variables is a generalization of \eqref{E123} that can be obtained using the derivatives of the moment-generating function:  
 \begin{eqnarray}
\mathbb{E}[X_1X_2X_3X_4]&=&\frac{\partial^4}{\partial t_1 \partial t_2 \partial t_3\partial t_4} \mathbb{E}[e^{\mathbf{t}^{\intercal}\bm{X}}] |_{t_{1,2,3,4}=0 }\nonumber \\
&=&  \frac{\partial^4}{\partial t_1 \partial t_2 \partial t_3 \partial t_4}  e^{\mathbf{t}^{\intercal}\bm{\mu}+\frac{1}{2}\mathbf{t}^{\intercal}\mathbf{\Sigma} \mathbf{t}}|_{t_{1,2,3,4}=0} \nonumber \\
&& \!\!\!\!\!\!\!\!\!\!\!\!\!\!\!\!\!\!\!\!\!\!\!\!\!\!\!\!\!\!\!\!\!\!\!\!=\text{cov}(X_1X_2)\big(\text{cov}(X_3,X_4)+\mu_3\mu_4\big)\nonumber \\
&& \!\!\!\!\!\!\!\!\!\!\!\!\!\!\!\!\!\!\!\!\!\!\!\!\!\!\!\!\!\!\!\!\!\!\!\! +\text{cov}(X_1X_3)\big(\text{cov}(X_2,X_4)+\mu_2\mu_4\big)\nonumber\\
&& \!\!\!\!\!\!\!\!\!\!\!\!\!\!\!\!\!\!\!\!\!\!\!\!\!\!\!\!\!\!\!\!\!\!\!\!+\text{cov}(X_2X_3)\big(\text{cov}(X_1,X_4)\nonumber \\
&& \!\!\!\!\!\!\!\!\!\!\!\!\!\!\!\!\!\!\!\!\!\!\!\!\!\!\!\!\!\!\!\!\!\!\!\! +\mu_1\mu_4\big)+\text{cov}(X_1,X_4)\mu_2\mu_3+\text{cov}(X_2,X_4)\mu_1\mu_3\nonumber\\
&& \!\!\!\!\!\!\!\!\!\!\!\!\!\!\!\!\!\!\!\!\!\!\!\!\!\!\!\!\!\!\!\!\!\!\!\!+\text{cov}(X_3,X_4)\mu_1\mu_2+\mu_1\mu_2\mu_3\mu_4. \nonumber 
\end{eqnarray}
Using the definition of variance
\begin{eqnarray}
\text{var}(X_1X_2) =\mathbb{E}[(X_1X_2)^2]-\mathbb{E}[X_1X_2]^2  \label{eq_v12}
\end{eqnarray}
The elements of variance can be expended as below 
\begin{eqnarray}
\mathbb{E}[(X_1X_2)^2]&=&\frac{\partial^4}{\partial t_1^2 \partial t_2^2} \mathbb{E}[e^{\mathbf{t}^{\intercal}\bm{X}}] |_{t_1 =t_2=t_3=t_4=0 } \nonumber\\
&=& \frac{\partial^4}{\partial t^2_1 \partial t^2_2}  e^{\mathbf{t}^{\intercal}\bm{\mu}+\frac{1}{2}\mathbf{t}^{\intercal}\mathbf{\Sigma} \mathbf{t}}|_{t_1 =t_2=t_3=t_4=0}\nonumber\\
&=&\sigma_{1}^{2}\sigma_{2}^{2}+2\text{cov}(X_1,X_2)^{2}\nonumber\\
&&+4\text{cov}(X_1,X_2)\mu_{1}\mu_{2}\nonumber\\
&&+\sigma_{1}^{2}\mu_{2}^{2}+\sigma_{2}^{2}\mu_{1}^{2}+\mu_{1}^2\mu_{2}^2. \label{eq_e12a}
\end{eqnarray}
\begin{eqnarray}
\mathbb{E}[X_1X_2]^2&=& \big(\text{cov}(X_1,X_2)+\mathbb{E}(X_1)\mathbb{E}(X_2)\big)^2\nonumber\\
&=&\text{cov}(X_1,X_2)^2+2\text{cov}(X_1,X_2)\mu_1\mu_2\nonumber\\
&&+\mu_1^2\mu_2^2. \label{eq_e12b}
\end{eqnarray}
Substituting \eqref{eq_e12a} and \eqref{eq_e12b} in \eqref{eq_v12} establishes \eqref{eq4}.


\section{Formulation for the binary tree hierarchical decomposition}\label{A:binary_tree}
\begin{figure}[t!]\centering
\includegraphics[width=0.35 \columnwidth]{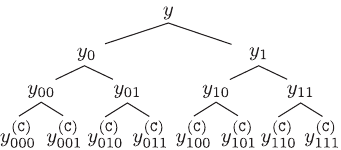}
\caption{Representation of a 3-layers hierarchical binary decomposition of classes $y_{ijk}^{(\mathtt{C})}\in\{1,2,\cdots,8\}$ using the output layer variables $\bm{y}=[y\,y_{0}\,y_{1}\,y_{00}\,y_{01}\,y_{10}\,y_{11}]^{\intercal}\in\mathbb{R}^{7}$.}
\label{FIG:Hierarchical_tree}
\end{figure}

For classification problem, each class is encoded in a binary tree with $\mathtt{H}=\lceil\log_2(\mathtt{K})\rceil$ layers, and which is defined by $\mathtt{Y}=\mathtt{K}-1$ hidden states when $\log_2(\mathtt{K})\in\mathbb{Z}^+$. Figure \ref{FIG:Hierarchical_tree} depicts the hierarchical decomposition for $\mathtt{K}=8$ classes and $\mathtt{H}=3$ layers, where a given class $y^{(\mathtt{C})}_{\mathcal{C}}$ such that $\mathcal{C}=\{j,k,l\}\in\{0,1\}^3$ is uniquely described by a set of $\mathtt{H}$ indices. In a binary context where $\mathtt{H}=1$, we can transform a regression problem into a probability for a class $y^{\mathtt{C}}_i,~i\in\{0,1\}$ by using 
$$p(y^{\mathtt{C}}_i|y)=\Phi\!\left(\!(-1)^i \frac{y}{\alpha}\right),$$
where, $\Phi(\cdot)$ denotes the standard normal CDF, and $\alpha\in\mathbb{R}^+$ is a scaling factor to account for the fact that the standard normal CDF reaches values close to 0 or 1 for inputs close to -3 and 3. Therefore, if the data is initially normalized in range -1 to 1, we need the scaling factor $\alpha\approx1/3$ in order to accommodate for this discrepancy. In the general case with $\mathtt{K}$-classes, the conditional probability of a class given the output values $\bm{y}$ is defined by  
$$p(y^{(\mathtt{C})}_{\mathcal{C}}|\bm{y})=\prod_{h=1}^{\mathtt{H}}\Phi\!\left(\!(-1)^{\mathcal{C}_h} \frac{[\bm{y}_\mathcal{C}]_{h}}{\alpha}\!\right),$$
where $\bm{y}_{\mathcal{C}}=[y\,y_{j}\,y_{jk}\,y_{jkl}\,\cdots]^\intercal\in \mathbb{R}^{\mathtt{H}}$. For the example in Figure \ref{FIG:Hierarchical_tree} where $\mathtt{H}=3$ layers, it simplifies to $\bm{y}_{\mathcal{C}}=[y\,y_{j}\,y_{jk}]^\intercal$, so that $$p(y^{(\mathtt{C})}_{\{ijk\}}|\bm{y})=\Phi\!\left(\!(-1)^i \frac{y}{\alpha}\!\right)\cdot \Phi\!\left(\!(-1)^j\frac{y_j}{\alpha}\!\right)\cdot \Phi\!\left(\!(-1)^k \frac{y_{jk}}{\alpha}\!\right).$$

For the special case where $\bm{Y}_{\mathcal{C}}|\mathcal{D}\sim\mathcal{N}(\bm{\mu}_{\bm{Y}_{\mathcal{C}}},\text{diag}(\bm{\sigma}_{\bm{Y}_{\mathcal{C}}}^2))$ follows a Gaussian distribution with diagonal covariance, we can employ the development found in \citeauthor{williams2006gaussian} (\citeyear{williams2006gaussian}) in order to obtain a closed-form solution to marginalize the output layer's uncertainty,
\begin{equation}\begin{array}{ll}
p(y^{(\mathtt{C})}_{\mathcal{C}}|\mathcal{D})&=\displaystyle\int p(y^{(\mathtt{C})}_{\mathcal{C}}|\bm{y}_{\mathcal{C}})\cdot f(\bm{y}_{\mathcal{C}}|\mathcal{D})d\bm{y}\\[-1pt]
&\displaystyle=\prod_{h=1}^{\mathtt{H}}\Phi\!\left(\!(-1)^{\mathcal{C}_h}\frac{[\bm{\mu}_{\bm{Y}_{\mathcal{C}}}]_h}{\sqrt{\alpha^2+[\bm{\sigma}_{\bm{Y}_{\mathcal{C}}}^2]_h}}\!\right)
\end{array}\label{EQ:class_marginal}\end{equation}
Note that in the case where the number of classes $\mathtt{K}$ does not correspond to an integer to the power $2$, it is required to normalize the marginal probabilities obtained in Equation \ref{EQ:class_marginal} in order to account for the unused leaves from the binary tree. During the training phase where we infer the network's parameters from observations $y^{(\mathtt{C})}\in\{1,2,\cdots,\mathtt{K}\}$, we convert each class into a $\mathtt{H}$-component vector $\bm{y}_{\mathcal{C}}\in\{-1,1\}^{\mathtt{H}}$ so that $[\bm{y}_{\mathcal{C}}]_i=(-1)^{\mathcal{C}_i}$. 

\section{Example of $\mathbf{F}$ matrices for product-terms indices}\label{Appendix:B}
\begin{figure}[h!]
\centering
\tikzset{dist1/.style={path picture= {
    \begin{scope}[x=1pt,y=10pt]
      \draw plot[domain=-6:6] (\x,{1/(1 + exp(-\x))-0.5});
    \end{scope}
    }
  }
}
\tikzset{dist2/.style={path picture= {
    \begin{scope}[x=1pt,y=10pt]
      \draw plot[domain=-6:6] (\x,{\x/10});
    \end{scope}
    }
  }
}
\tikzstyle{input}=[draw,fill=red!10,circle,minimum size=15pt,inner sep=0pt]
\tikzstyle{activ}=[draw,fill=cyan!20,circle,minimum size=15pt,inner sep=0pt]
\tikzstyle{hidden}=[draw,fill=green!20,circle,minimum size=15pt,inner sep=0pt]
\tikzstyle{output}=[draw,fill=blue!20,circle,minimum size=15pt,inner sep=0pt]
\tikzstyle{bias}=[draw=none,circle,minimum size=15pt,inner sep=0pt]

\tikzstyle{stateTransition}=[->,line width=0.25pt,draw=black!50]

\begin{tikzpicture}[scale=1.36]\scriptsize
    \node (h11)[activ] at (1, 1.5) {$a^{(1)}_{1}$};
    \node (h12)[activ] at (1, 0.5) {$a^{(1)}_{2}$};
    \node (b1)[bias]  at (1.35,1.9) {$\bm{b}^{(1)}$};

    \node (h21)[hidden] at (2, 1.5) {$z^{(2)}_{1}$};
    \node (h22)[hidden] at (2, 0.5) {$z^{(2)}_{2}$};

    \draw[stateTransition,opacity=1] (h11) -- (h21) node [midway, rotate=0,fill=white,opacity=0.8] {\scriptsize$w^{(1)}_{1,1}$};
    \draw[stateTransition,opacity=0] (h11) -- (h21) node [midway, rotate=0,fill=white] {\scriptsize$w^{(1)}_{1,1}$};
    \draw[stateTransition,opacity=0.5] (h11) -- (h22) node  {};
  
    \draw[stateTransition,opacity=1] (h12) -- (h21) node [midway, rotate=45,fill=white,opacity=0.8] {\scriptsize$w^{(1)}_{1,2}$};
        \draw[stateTransition,opacity=0] (h12) -- (h21) node [midway, rotate=45,fill=white!90] {\scriptsize$w^{(1)}_{2,1}$};
    \draw[stateTransition,opacity=0.5] (h12) -- (h22) node  {};
   
       \draw[stateTransition,opacity=1] (b1) -- (h21) node  [midway,above=-0.05cm, rotate=-25] {\scriptsize$b^{(1)}_{1}$};
    \draw[stateTransition,opacity=0.5] (b1) -- (h22) node {};


\end{tikzpicture}
\caption{Example of trivial network configuration employed to illustrate the configuration for the matrices $\mathbf{F}_{\bm{wa}}^{(j)}$ and $\mathbf{F}_{\bm{b}}^{(j)}$.}
\label{FIG:FZ}
\end{figure}
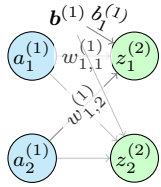

Figure \ref{FIG:FZ} presents an example of two successive hidden layers each comprising only two hidden units. 
The formulation of the the $\mathbf{F}_{\bm{wa}}^{(j)}$ and in $\mathbf{F}_{\bm{b}}^{(j)}$ matrices corresponding to the network in Figure \ref{FIG:FZ} is
$$\underbrace{\small\begin{bmatrix}z^{(2)}_1\\[4pt]z^{(2)}_2\end{bmatrix}}_{\bm{z}^{(j+1)}}=\underbrace{\small\begin{bmatrix}1&1&0&0\\0&0&1&1\end{bmatrix}}_{\mathbf{F}_{\bm{wa}}^{(j)}}\times \underbrace{\small\begin{bmatrix}w^{(1)}_{1,1}a^{(1)}_1\\[4pt]w^{(1)}_{1,2}a^{(1)}_2\\[4pt]w^{(1)}_{2,1}a^{(1)}_1\\[4pt]w^{(1)}_{2,2}a^{(1)}_2\end{bmatrix}}_{(\bm{w\!\!a})^{(j)}}+\underbrace{\small\begin{bmatrix}1&0\\0&1\end{bmatrix}}_{\mathbf{F}_{\bm{b}}^{(j)}}\times\underbrace{\small\begin{bmatrix}b^{(1)}_1\\[4pt]b^{(1)}_2\end{bmatrix}}_{\bm{b}^{(j)}}.$$
Note that the structure of $\mathbf{F}_{\bm{wa}}^{(j)}$ depends on the ordering of variables.

\section{Experiment configurations and supplementary results for benchmark regression datasets}\label{Appendix:C}
Table \ref{TAB:benchmark} presents the details for the experiments conducted for the benchmark regression datasets. Note that the times presented in the last columns are the average parameter (i.e. weights and biases) inference time in second per folds and the average hyper parameter (i.e. $\sigma_V$) optimization time in second. All these experiments were conducted using CPU.
\def\arraystretch{1}
\setlength\tabcolsep{1mm}
\begin{table*}[htbp]\small
\centering
\caption{Experiment details for the benchmark regression datasets using BLNN. $\mathtt{X}$: number of covariates, $\mathtt{L}$: number of layers, $\mathtt{A}$: number of activation units per layer, $\mathtt{F}$ number of random training/test folds. }
\scalebox{1}{\begin{tabular}{r|lllllllll}
\toprule
&&{Train}&{Test} &&& Average inference ($\bm{\theta}$)&Average optimization ($\sigma_V$)&Average $\sigma_V$ \\[4pt]
Datasets&$\mathtt{X}$& \#obs.&\#obs.&$\mathtt{L}\times\mathtt{A}$&$\mathtt{F}$&time per fold (s)&time per fold (s)&(20 folds)\\
\cmidrule{1-9}
Boston 	&13&455&51&1$\times$50&20&0.6&1.5&0.28\\
Concrete 			&8&927&103&1$\times$50&20&0.9&2.3&0.32\\
Energy 			&8&691&77&1$\times$50&20&0.6&2.7&0.15\\
Kin8nm 			&8&7373&819&1$\times$50&20&6.7&16.7&0.36\\
Naval          &16&11934&1193&1$\times$50&20&13.7&32.6&0.31\\
Power 		&4&8611&957&1$\times$50&20&6.6&17.7&0.24\\
Protein 			&9&41157&4373&1$\times$100&5&39&53.5&0.74\\
Wine 			&11&1439&160&1$\times$50&20&1.4&2.1&0.72\\
Yacht 			&6&277&31&1$\times$50&20&0.2&1.4&0.07\\
\bottomrule
\end{tabular}}
\label{TAB:benchmark}
\end{table*}

\end{document}